\documentclass[twoside,leqno,twocolumn]{article}

\usepackage[letterpaper]{geometry}
\usepackage[colorlinks]{hyperref}
\hypersetup{
  colorlinks = true,
  linkcolor = black,   
  citecolor = green, 
  urlcolor = cyan,    
  anchorcolor = black 
}

\usepackage{ltexpprt}
\usepackage{hyperref}
\usepackage{multirow}
\usepackage{cite}
\usepackage{amsmath,amssymb,amsfonts}
\usepackage{algorithmic}
\usepackage{graphicx}
\usepackage{textcomp}
\usepackage{xcolor}
\usepackage{booktabs}
\usepackage{subfigure}
\usepackage{pgfplots}
\def\sysname{DDMC}

\begin{document}

\newcommand\relatedversion{}
\renewcommand\relatedversion{\thanks{The full version of the paper can be accessed at \protect\url{https://arxiv.org/abs/1902.09310}}} 

\title{\Large Dual-disentangled Deep Multiple Clustering}
\author{Jiawei Yao\thanks{\scriptsize{\{jwyao, juhuah\}@uw.edu, School of Engineering and Technology, University of Washington, Tacoma, WA 98402, USA}}
\and Juhua Hu\footnotemark[1]} 


\date{}

\maketitle







\begin{abstract} \small\baselineskip=9pt Multiple clustering has gathered significant attention in recent years due to its potential to reveal multiple hidden structures of the data from different perspectives. 
Most of multiple clustering methods first derive feature representations by controlling the dissimilarity among them, subsequently employing traditional clustering methods (e.g., k-means) to achieve the final multiple clustering outcomes. However, the learned feature representations can exhibit a weak relevance to the ultimate goal of distinct clustering. Moreover, these features are often not explicitly learned for the purpose of clustering. Therefore, in this paper, we propose a novel Dual-Disentangled deep Multiple Clustering method named \sysname{} by learning disentangled representations. 
Specifically, \sysname{} is achieved by a variational Expectation-Maximization (EM) framework. 
In the E-step, the disentanglement learning module employs coarse-grained and fine-grained disentangled representations to obtain a more diverse set of latent factors from the data.
In the M-step, the cluster assignment module utilizes a cluster objective function to augment the effectiveness of the cluster output. Our extensive experiments demonstrate that \sysname{} consistently outperforms state-of-the-art methods across seven commonly used tasks. Our code is available at \href{https://github.com/Alexander-Yao/DDMC}{https://github.com/Alexander-Yao/DDMC}.
\end{abstract}

\section{Introduction}
\begin{figure}
    \centering
    \includegraphics[width=0.45\textwidth]{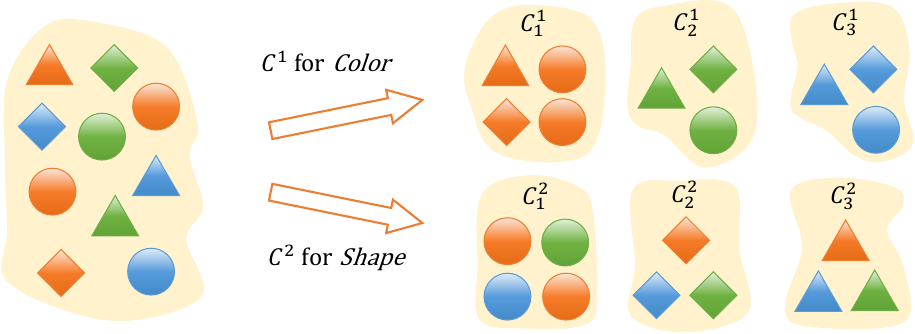}
    \caption{\textbf{An example of multiple clustering }that can reveal two or more distinct clusterings (i.e., $C^1$ for color and $C^2$ for shape).}
    \label{fig:intro}
\end{figure}

Clustering, which groups data points based on their similarities, has been extensively researched, since huge amount of unlabeled data are becoming more and more available.
Traditional methods like k-means~\cite{macqueen1967some}, spectral clustering~\cite{ng2001spectral}, and Gaussian mixture model~\cite{bishop2006pattern}, use general-purpose handcrafted features that are not always ideal for specific tasks. Recently, 
deep clustering algorithms~\cite{xie2016unsupervised,guerin2018improving, qian2022unsupervised} leveraging Deep Neural Networks (DNNs) significantly improve performance. Whereas most algorithms yield a single data partition, multiple clustering algorithms have been developed to generate different partitions for varying applications, demonstrating the ability to identify multiple distinct clusterings from a dataset, as shown in Fig.~\ref{fig:intro}.

Existing multiple clustering methods can be roughly divided into shallow and deep models.
For shallow models, COALA~\cite{bae2006coala} treats objects within an established clustering as constraints for generating an alternative clustering. Some other ones are relying on different feature subspaces, e.g., Hu et al.~\cite{hu2017finding} proved the relation between Laplacian eigengap and stability of a clustering, and discovered multiple clusterings via maximizing the eigengap within different feature subspaces.
Recent developments have observed researchers employing deep learning to generate multiple clusterings. ENRC~\cite{miklautz2020deep} merges an auto-encoder and clustering objective function to yield alternative clusterings. iMClusts~\cite{ren2022diversified} harnesses the representational capabilities of deep autoencoders and multi-head attention to generate multiple clusterings. AugDMC~\cite{DBLP:conf/inns-dlia/YaoLRH23} leverages augmentation to learn distinct representations for multiple clusterings.


Although deep multiple clustering methods have yielded impressive outcomes, they still confront two main challenges. Firstly, the relevance between the learned representations and the ultimate goal of distinct clusterings is weak. This issue arises because the diversity of clusterings is indirectly enforced by limiting the overlap of learned representations. However, this does not guarantee a direct correlation between the dissimilarity of feature representations and the clustering diversity, potentially leading to redundant clusterings. Secondly, most existing methods simply feed the learned representations into traditional clustering algorithms, such as k-means, to obtain multiple clusterings~\cite{ren2022diversified,DBLP:conf/inns-dlia/YaoLRH23}. However, the representations are often learned without involvement of the clustering objective, thereby undermining the final clustering outcomes. Although some efforts, like ENRC~\cite{miklautz2020deep}, have aimed to optimize the clustering performance, they have not yet achieved satisfactory results~\cite{ren2022diversified,DBLP:conf/inns-dlia/YaoLRH23}. 

Fortunately, disentangled representation learning, aiming to learn factorized representations that uncover and separate the latent factors hidden in data~\cite{he2017unsupervised,bao2019generating,cheng2020improving}, can help learn the diverse representations effectively for multiple clustering. Considering Fig.~\ref{fig:intro} as an example, the objects have at least two distinct factors (i.e., shape and color). Disentangled representation learning can segregate these factors and encode them into independent and distinct latent variables within the representation space~\cite{higgins2017beta}. Consequently, the latent variable of shape/color changes exclusively with the variation of the object's shape/color and remains constant relative to other factors. 
Despite the success of disentangled representation learning, no study to date has examined its use in achieving multiple clustering.


However, the application of disentangled representation learning towards multiple clustering is not straightforward. 
First, 
disentangled representation learning, despite its remarkable success, was not initially designed for multiple clustering. Therefore, the design of a disentangled representation learning framework specifically intended for multiple clustering becomes crucial. Second, the effectiveness in purpose of clustering needs to be ensured. Most of existing deep multiple clustering methods emphasize primarily on capturing features at the clustering level, thereby neglecting the effectiveness at the cluster level within each individual clustering. 


Therefore, in this paper, we aim for a novel Dual-Disentangled deep Multiple Clustering (\sysname{}) method that can simultaneously ensure clustering-level and cluster-level performance through an end-to-end approach. 
Specifically, our disentanglement learning module leverages coarse-grained and fine-grained disentangled representations \textit{to learn more diverse disentangled representations}. 
Simultaneously, our cluster assignment module is designed \textit{to boost the effectiveness of the proposed method in cluster-level performance}. 
We structure our approach as a variational Expectation-Maximization (EM) framework. In the Expectation (E) step, we decipher unique disentangled representations to reveal potential multiple clusterings, while the cluster assignment component is fixed. During the Maximization (M) step, the disentangled representations obtained from the E-step are fixed and then leveraged in the cluster assignment learning process. 
The contributions of this work can be summarized as
\begin{itemize}
    \item We propose a novel dual-disentangled deep multiple clustering method, which is the first to introduce disentanglement learning for multiple clustering.
    \item Our proposal is achieved by a variational EM framework. In E-step, the disentangled representation is learned to enable the achievement of multiple clustering. In M-step, cluster assignment is optimized to enhance the cluster-level performance.
    \item We conducted extensive experiments on seven commonly used tasks, which demonstrates the superiority of our proposal \sysname{}. 
    
\end{itemize}

\begin{figure*} [t]
    \centering
    \includegraphics[width=0.83\textwidth]{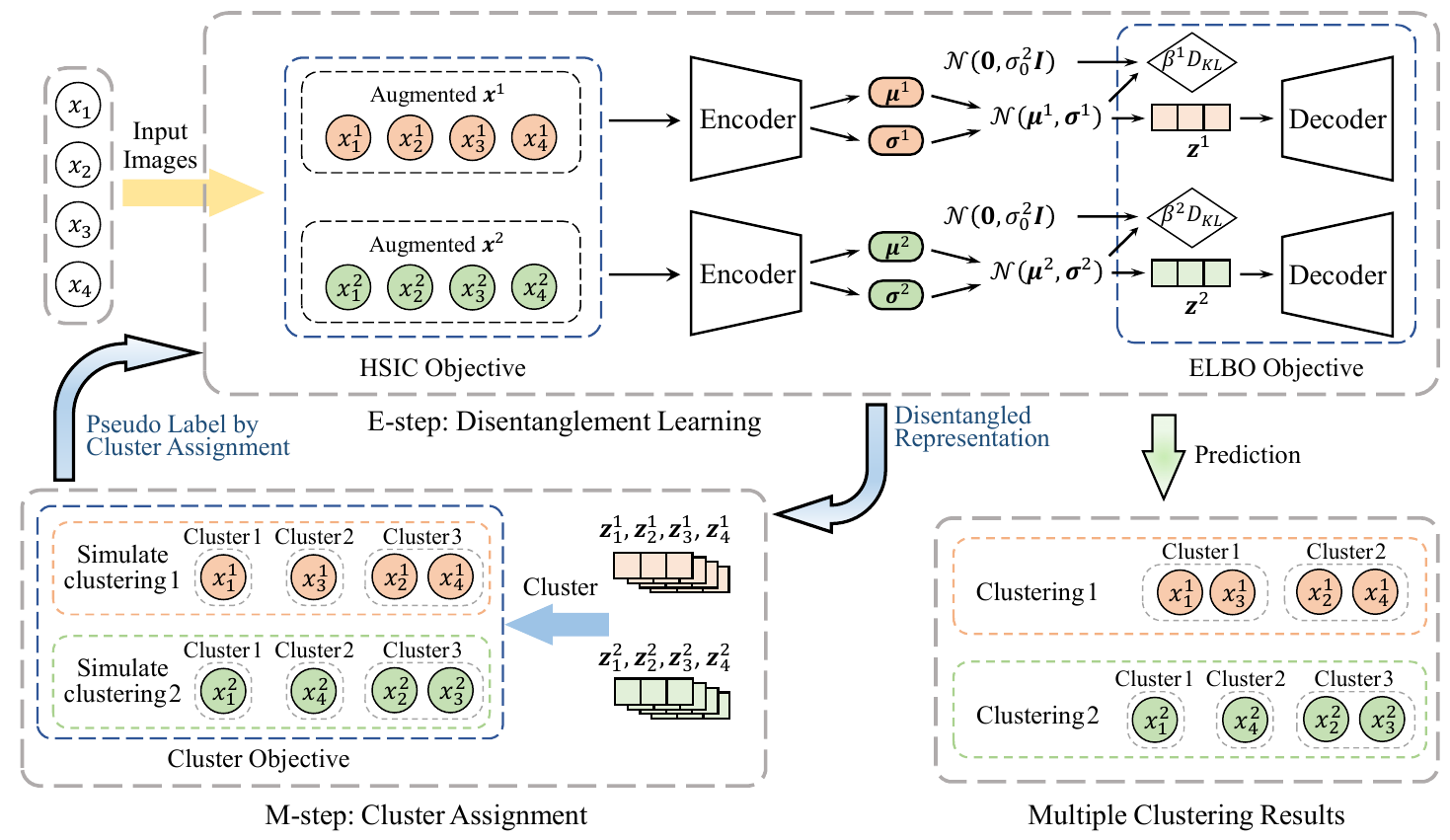}
    \caption{\textbf{\sysname{} framework} trains disentanglement learning and cluster assignment in an EM framework. During the E-step, the disentangled representation is learned through both coarse-grained and fine-grained disentangled representation learning. The learned disentangled representations can be applied to multiple clustering tasks. In the M-step, cluster assignment is optimized, enhancing the cluster-level performance.}
    \label{fig:model}
\end{figure*}
\section{Related Work}

\subsection{Multiple Clustering}
The data being analyzed could contain natural alternative perspectives. Multiple clustering, as a kind of method that can discover these alternative perspectives of data, has attracted considerable attention.
Traditional methods for multiple clustering~\cite{hu2018subspace} use shallow models to find different ways of grouping data. Some of these methods rely on constraints to produce alternative clusterings. For instance, COALA~\cite{bae2006coala} uses the objects in an existing clustering as constraints for creating a new clustering and Qi et al.~\cite{qi2009principled} formulated multiple clustering as a constrained optimization problem. 
Other methods exploit different feature subspaces to generate multiple clusterings. For example, Hu et al.~\cite{hu2017finding} found multiple clusterings by maximizing the eigengap in different subspaces. MNMF~\cite{yang2017non} adds the inner product of similarity matrices as a regularization term to find multiple clusterings. Some methods also use information theory to generate multiple clusterings. Gondek and Hofmann~\cite{gondek2003conditional} applied conditional information bottleneck and
Dang and Bailey~\cite{dang2010generation} used an expectation maximization framework to optimize mutual information. 

Recently, some methods have used deep learning to find multiple clusterings and achieved better results. Wei et al.~\cite{wei2020multi} proposed a deep matrix factorization based method that used multi-view data to find multiple clusterings. ENRC~\cite{miklautz2020deep} uses an auto-encoder to learn object features and finds multiple clusterings by optimizing a clustering objective function. iMClusts~\cite{ren2022diversified} exploits auto-encoders and multi-head attention to learn features from different perspectives and finds multiple clusterings. AugDMC~\cite{DBLP:conf/inns-dlia/YaoLRH23} leverages data augmentation to generate diverse images and learns the representations to discover multiple clusterings. 
However, these methods suffer from the inability to obtain feature representations closely relevant to the goal of distinct clusterings, and the learned features without a good involvement of the clustering objective can be unsuitable for sufficient cluster-level performance.

\subsection{Disentangled Representation Learning}
The goal of disentangled representation learning is to learn factorized representations that uncover and separate the latent factors hidden in data. 
In computer vision, Higgins et al.~\cite{higgins2017beta} proposed $\beta$-VAE to learn interpretable factorized latent representations from raw image data without supervision. InfoGAN~\cite{chen2016infogan} leverages GAN paradigm with an extra variational regularization of mutual information to learn disentangled representations. InfoSwap~\cite{gao2021information} disentangles identity-relevant and identity-irrelevant information through optimizing information bottleneck to generate more identity discriminative swapped faces. In natural language processing, Hu et al.~\cite{hu2017toward} combined VAE with an attribute discriminator to disentangle content and attributes of the given textual data, for generating texts with desired attributes of sentiment and tenses. Bao et al.~\cite{bao2019generating} generated sentences from disentangled syntactic and semantic spaces by modeling syntactic information in the latent space of VAE and regularizing syntactic and semantic spaces via an adversarial reconstruction loss.

Although these methods have achieved good results, there are rare studies on clustering scenarios. MultiVAE~\cite{xu2021multi} learns the disentangled representations from multi-view data.  However, the exploration of disentangled representations in the context of multiple clustering remains a seldom investigated area of research. 
\section{The Proposed Method}

To simultaneously learn representations for distinct clusterings and achieve good cluster-level performance, we leverage disentangled representation learning and cluster assignment learning within a variational EM framework 
as illustrated in Fig.~\ref{fig:model}. 
Specifically, given an image  $\boldsymbol{x}_i\in \{ \boldsymbol{x}_i \}^N_{i=1}$, our aim in disentangled multiple clustering is to derive $K$ distinct image representations $\{ \boldsymbol{z}_i^{1},\dots, \boldsymbol{z}_i^{K} \}$, which describe various facets of the image through achieving both coarse and fine-grained disentanglement. 
Consequently, these images can be classified into $M$ distinct clusterings, each reflecting a unique aspect of the original image, where $K$ can be larger than $M$. This is because that the real-world data may have more aspects than the desired number, while all the aspects have to be disentangled to obtain the needed representations.

An image can encapsulate diverse aspects, each of which can correspond to a clustering perspective. To effectively uncover these latent facets, we strive for coarse-grained disentanglement achieved through augmentation. Consequently, employing a variety of augmentation methods generates variant images that each mirror a distinct facet of the original image, thereby highlighting the inherent diversity as follows.

\subsection{The Coarse-grained Disentanglement}




This is achieved by sampling and mixing. In the sampling stage, augmentation operations are sampled from a defined set of augmentation operations. This sampling procedure is repeated $K$ times, each corresponding to a unique disentangled representation set. During the $k$-th sampling iteration, the chosen augmentation methods are applied to the image set $\boldsymbol{X}= \{ \boldsymbol{x}_i \}^N_{i=1}$ to produce the augmented images $\boldsymbol{X}_{aug}^k$.

A prevalent issue in deep learning is that applying multiple augmentations to an image can make its representation unstable if there is no label information~\cite{cubuk2020randaugment}. 
To alleviate this issue, we leverage the mixing process that generates a variety of transformations for robustness. Specifically, for the $k$-th coarse-grained disentanglement, the augmented images can be generated by
\begin{equation}
    \boldsymbol{X}^k = w^k\boldsymbol{X} + (1-w^k) \boldsymbol{X}^k_{aug}
\end{equation}
where $w^k\in(0, 1 )$ is the mixing weight.
To further distinguish the $K$ augmented image sets, we set $w^k$ as trainable parameters and use Hilbert Schmidt Independence Criterion (HSIC)~\cite{gretton2005measuring} to measure the dependency between two augmented image sets $\boldsymbol{X}^k$ and $\boldsymbol{X}^{k'}$ as
\begin{equation}
    \text{HSIC}(\boldsymbol{X}^k,\boldsymbol{X}^{k'})=(N-1)^2\operatorname{tr}\big(\boldsymbol{G}_k\boldsymbol{H}\boldsymbol{G}_{k'}\boldsymbol{H}\big)
\end{equation}
where $\operatorname{tr}(\cdot)$ is the matrix trace operator and $\boldsymbol{H}=\boldsymbol{I}_N-\frac{1}{N} \boldsymbol{1}\boldsymbol{1}^{\top}$ is the centering matrix.
$\boldsymbol{G}_k\in \mathbb{R}^{N\times N}$ is the kernel matrix to measure the similarity within the image set $\boldsymbol{X}^k$. 
Here we adopt the inner product kernel for simplicity as $\boldsymbol{G}_k=(\boldsymbol{X}^k)^{\top}\boldsymbol{X}^k$. A smaller HSIC value signifies greater independence between the two augmented image sets. Therefore, we can maximize the negative HSIC value across the $K$ augmented image sets to learn a more diverse augmentation manner to do the coarse-grained disentanglement as
\begin{equation}\small\nonumber
\begin{aligned}
\mathcal{L}_{\text{coarse}} = &
-\sum_{k=1, k \neq k^{\prime}}^K \text{HSIC}\left(\boldsymbol{X}^k, \boldsymbol{X}^{k^{\prime}}\right)\\
=&- (N-1)^2 \sum_{k=1, k \neq k^{'}}^K \operatorname{tr}\left(\boldsymbol{X}_k^{\top} \boldsymbol{X}_k \boldsymbol{H} (\boldsymbol{X}^{k'})^{\top} \boldsymbol{X}^{k'} \boldsymbol{H} \right)
\end{aligned}
\end{equation}



\subsection{The Fine-grained Disentanglement}
Although the coarse-grained disentanglement helps explore the diversity of images, they may not be sufficient to obtain effective image representations for multiple clustering. Therefore, we propose fine-grained disentanglement to decipher the factorized representation of the augmented images, denoted as $\{\boldsymbol{z}_i^{k}\}_{k=1}^{K}$ with $\boldsymbol{z}_i^{k}$ indicating the $k$-th high-level aspect for image $\boldsymbol{x}_i$. Additionally, we infer a set of one-hot vectors, $\boldsymbol{C}=\{\boldsymbol{c}^k\}_{k=1}^K$, for each high-level aspect. 
Assuming the images can be grouped into $M$ clusterings, the dimension of $\boldsymbol{c}^k$ is equal to the number of clusterings, i.e., $\boldsymbol{c}^k=[c_{1}^k, c_{2}^k, \dots, c_{M}^k]$. If the $k$-th disentangled representation falls under the clustering $m$, $c_m^k=1$ and $c_{m'}^{k}=0$ for all $m'\neq m$, 
ensuring each representation is clearly associated with a single clustering.

Each element $c_m^k\in\boldsymbol{c}^k$ is described by a distribution $p(c_m^k)$, which denotes the probability of the disentangled representation $\boldsymbol{Z}^k=\{\boldsymbol{z}_i^k\}_{i=1}^N$ corresponding to the $m$-th clustering.
Thereafter, we consider the joint probability:
\begin{equation}
\begin{split}
    p(\boldsymbol{X}^k, \boldsymbol{Z}^k, \boldsymbol{c}^k) & = p(\boldsymbol{X}^k|\boldsymbol{Z}^k, \boldsymbol{c}^k) p(\boldsymbol{Z}^k, \boldsymbol{c}^k) \\
    & = p(\boldsymbol{X}^k|\boldsymbol{Z}^k, \boldsymbol{c}^k) p(\boldsymbol{Z}^k) p( \boldsymbol{c}^k)
\end{split}
\end{equation}
Given that both $\boldsymbol{c}^k$ and $\boldsymbol{Z}^k$ are dependent on the augmented images $\boldsymbol{X}^k$, 
the posterior of $\boldsymbol{c}^k$ and $\boldsymbol{Z}^k$ are written as $p(\boldsymbol{c}^k|\boldsymbol{X}^k)$ and $p(\boldsymbol{Z}^k|\boldsymbol{X}^k)$, respectively.
The integral of the posterior in VAE is intractable, so we employ $q_{\phi^k}(\boldsymbol{c}^k|\boldsymbol{X}^k)$ and $q_{\phi^k}(\boldsymbol{Z}^k|\boldsymbol{X}^k)$, which are parameterized by $\phi^k$, as approximations for the true posterior.
Since the aspect assignment $\boldsymbol{c}^k$ is a one-hot representation, which is non-differentiable during the training process, we set its prior as a product of independent uniform Gumbel Softmax distributions, i.e., $p(\boldsymbol{c}^k)=p(c_1^k)p(c_2^k)\dots p(c_M^k)$, where $p(c_m^k) \sim Gumbel(0,1)$. Hence, the approximate posterior $q_{\phi^k}(\boldsymbol{c}^k|\boldsymbol{X}^k)$ becomes
\begin{equation}
\begin{split}
    q_{\phi ^k} (\boldsymbol{c}^k|\boldsymbol{X}^k)  = & \prod_{m=1}^M q_{\phi^k}(c^k_m|\boldsymbol{X}^k)\\
    = & \prod_{m=1}^M \frac{\exp\big( (\log s_k + g_m^k) / \tau \big)}{\sum_{i=1}^K \exp\big( (\log s_i + g_m^k) / \tau \big)}
\end{split}
\end{equation}
where $g_m^k\sim Gumbel(0,1)$ and $\tau$ is the temperature parameter exploited to control the scale of values. The shared trainable parameter $s_k\in\{ s_1,s_2,\dots,s_K \}$ is employed to generate all the aspect assignment parameters $\{\boldsymbol{c}^1, \dots, \boldsymbol{c}^K\}$, in a principle manner.
Conversely, the posterior $q_{\phi^k}(\boldsymbol{Z}^k|\boldsymbol{X}^k)$ can be parameterized by a factorized Gaussian as
\begin{equation}
    q_{\phi^k}(\boldsymbol{Z}^k|\boldsymbol{X}^k) = \prod_{i=1}^{N} q_{\phi^k}(\boldsymbol{z}_i^k|\boldsymbol{X}^k)
\end{equation}
Assuming that the prior of the disentangled representation $\boldsymbol{Z}^k$ follows a normal distribution, that is, $p(\boldsymbol{Z}^k)\sim \mathcal{N}(\boldsymbol{0}, (\sigma_0^k)^2 \mathbf{I})$. According to reparameterization trick~\cite{kingma2013auto}, the encoder $q_{\phi^k}(\boldsymbol{z}_i^k|\boldsymbol{X}^k)= \mathcal{N}(\boldsymbol{\mu}^k, {(\boldsymbol{\sigma}^k})^2)$ can be computed by a neural network $f_{nn}: \mathbb{R}^d\rightarrow\mathbb{R}^{2d}$
\begin{equation}\nonumber
    \boldsymbol{a}^k, \boldsymbol{b}^k = f_{nn}(\boldsymbol{x}^k),\,
    \boldsymbol{\mu}^k = \boldsymbol{a}^k, \,
    \boldsymbol{\sigma}^k\leftarrow \sigma_0^k \cdot \exp(-\frac{1}{2} \boldsymbol{b}^k)
\end{equation}
The neural network $f_{nn}(\cdot)$ 
is shared across the $K$ disentangled representations. In real tasks, we typically set $\sigma^{k}$  to a relatively small value like $0.2$ to prevent the inferred values from becoming excessively large.



The objective of fine-grained disentanglement is to maximize the likelihood of the augmented images. According to Jensen's inequality, the log-likelihood of our proposed model is
\begin{equation}\small\nonumber
\begin{split}
    \sum_{k=1}^K \log p(\boldsymbol{X}^k) =& \sum_{k=1}^K\log \int_{\boldsymbol{Z}^k}\sum_{\boldsymbol{c}} p(\boldsymbol{X}^k, \boldsymbol{Z}^k, \boldsymbol{c}^k) \operatorname{d}\!\boldsymbol{Z}^k \\
    \geq & \sum_{k=1}^K\mathbb{E}_{q(\boldsymbol{Z}^k, \boldsymbol{c}^k | \boldsymbol{X}^k)}\Bigg[ \log \frac{p(\boldsymbol{X}^k, \boldsymbol{Z}^k, \boldsymbol{c}^k)}{q(\boldsymbol{Z}^k, \boldsymbol{c}^k|\boldsymbol{X}^k)} \Bigg] \\
    = & \sum_{k=1}^K \mathcal{L}_{\text{ELBO}}(\boldsymbol{X}^k)
\end{split}
\end{equation}
where $\mathcal{L}_{\text{ELBO}}(\boldsymbol{X}^k)$ is the Evidence Lower Bound (ELBO) of the $k$-th disentangled representation. Maximizing the ELBO is equivalent to maximizing the likelihood in variational inference process. Given $p(\boldsymbol{X}^k, \boldsymbol{Z}^k, \boldsymbol{c}^k) = p(\boldsymbol{X}^k|\boldsymbol{Z}^k, \boldsymbol{c}^k)p(\boldsymbol{Z}^k, \boldsymbol{c}^k)$, ELBO of the $k$-th disentangled representation can be formulated as
\begin{equation}\label{eq:elbo1}\small
\begin{split}
    \mathcal{L}_{\text{ELBO}} (\boldsymbol{X}^k) & = \mathbb{E}_{q(\boldsymbol{Z}^k, \boldsymbol{c}^k|\boldsymbol{X}^k)}[\log p(\boldsymbol{X}^k|\boldsymbol{Z}^k, \boldsymbol{c}^k)]\\
    & - D_{KL}(q(\boldsymbol{Z}^k, \boldsymbol{c}^k | \boldsymbol{X}^k) \| p(\boldsymbol{Z}^k, \boldsymbol{c}^k))
\end{split}
\end{equation}
Assuming that the aspect assignment and disentangled representation are conditionally independent, i.e., $q(\boldsymbol{Z}^k, \boldsymbol{c}^k|\boldsymbol{X}^k) = q(\boldsymbol{Z}^k|\boldsymbol{X}^k)q(\boldsymbol{c}^k|\boldsymbol{X}^k)$ and the prior $p(\boldsymbol{Z}^k, \boldsymbol{c}^k)=p(\boldsymbol{Z}^k)p(\boldsymbol{c}^k)$.
The KL divergence can be factored into $D_{KL}( q(\boldsymbol{Z}^k|\boldsymbol{X}^k) \| p(\boldsymbol{Z}^k) )$ and $D_{KL}( q(\boldsymbol{c}^k|\boldsymbol{X}^k) \| p(\boldsymbol{c}^k) )$.
%
Thereafter, the KL divergence terms of $\boldsymbol{c}^k$ and $\boldsymbol{z}^k$ are separated, corresponding to aspect assignment and disentangled representations, respectively. Thereafter, for the $k$-th disentangled representation, the objective becomes to maximize
\begin{equation}\label{eq:elbo2}\small
\begin{split}
    \mathcal{L}_{ELBO}&(\boldsymbol{X}^k) = \mathbb{E}_{q(\boldsymbol{Z}^k, \boldsymbol{c}^k|\boldsymbol{X}^k)}[\log p(\boldsymbol{X}^k|\boldsymbol{Z}^k, \boldsymbol{c}^k)] \\- &
    D_{KL}\big( q(\boldsymbol{Z}^k|\boldsymbol{X}^k) \| p(\boldsymbol{Z}^k) \big) 
    - D_{KL}\big( q(\boldsymbol{c}^k|\boldsymbol{X}^k) \| p(\boldsymbol{c}^k) \big)
\end{split}
\end{equation}

\subsection{E-step: Disentanglement Optimization}
Optimizing the disentangled representations primarily involves fine-tuning the fine-grained disentanglement.
To capture more variation information,
the channel capacity of the KL divergence terms in Eqn.\eqref{eq:elbo2} should increase gradually.
We define controlled capacities $U_c$ and $U_z$ for the KL divergence terms of the aspect assignment variable $\boldsymbol{c}^k$ and the disentangled representation $\boldsymbol{Z}^k$ for the $k$-th disentangled representation, respectively. The, the ELBO can be reformulated as
\begin{equation}\nonumber
\begin{split}
    \mathcal{L}_{ELBO}(\boldsymbol{X}^k) =& \mathbb{E}_{q(\boldsymbol{Z}^k, \boldsymbol{c}^k|\boldsymbol{X}^k)}[\log p(\boldsymbol{X}^k|\boldsymbol{Z}^k, \boldsymbol{c}^k)] \\
    - &
    \beta^k|D_{KL}\big( q(\boldsymbol{Z}^k|\boldsymbol{X}^k) \| p(\boldsymbol{Z}^k) \big) 
 - U_z|\\
    - &\beta^k | D_{KL}\big( q(\boldsymbol{c}^k|\boldsymbol{X}^k) \| p(\boldsymbol{c}^k) \big) - U_c|
\end{split}
\end{equation}
where $\beta^k$ is a trade-off coefficient. 
Since different disentangled representation sets have different scale of data reconstruction loss, $\beta^k$ can be calculated as
\begin{equation}
    \beta^k = \beta\frac{\mathbb{E}_{q(\boldsymbol{Z}^k, \boldsymbol{c}^k|\boldsymbol{X}^k)} [\log p(\boldsymbol{X}^k|\boldsymbol{Z}^k, \boldsymbol{c}^k)]}{\max_{k}\mathbb{E}_{q(\boldsymbol{Z}^k\boldsymbol{c}^k|\boldsymbol{X}^k)} [\log p(\boldsymbol{X}^k|\boldsymbol{Z}^k, \boldsymbol{c}^k)]}
\end{equation}
Eventually, the fine-grained disentangled objective function can be written as
\begin{equation}\small\nonumber
    \begin{split}
        \mathcal{L}_{\text{fine}} =& \sum_{k=1}^{K} \mathcal{L}_{\text{ELBO}}(\boldsymbol{X}^k) \\
         =&\sum_{k=1}^K \mathbb{E}_{q\left(\boldsymbol{Z}^k, \boldsymbol{c}^k \mid\boldsymbol{X}^k\right)}\left[\log p\left(\boldsymbol{X}^k \mid \boldsymbol{Z}^k, \boldsymbol{c}^k\right)\right] \\
        & -\sum_{k=1}^K \beta^k\left|D_{K L}\left(q\left(\boldsymbol{Z}^k \mid \boldsymbol{X}^k\right)|| p\left(\boldsymbol{Z}^k\right)\right)-U_z\right| \\ 
        & - \sum_{k=1}^K \beta^k\left|D_{K L}\left(q\left(\boldsymbol{c}^k \mid\boldsymbol{X}^k\right)|| p(\boldsymbol{c}^k)\right)-U_c\right| 
    \end{split}
\end{equation}


\subsection{M-step: Cluster Assignment}
After the disentanglement is optimized, we can enhance the clustering performance by optimizing the following objective~\cite{yang2017towards}:
\begin{equation}\label{eq:loss_cluster}
\begin{split}
    \mathcal{L}_{\text {cluster}} =& - \max_{\mathbf{s}} \frac{1}{N} \sum_{i=1}^N\left\|\boldsymbol{z}
    _i^k-\mathbf{W}^k s_i\right\|_2^2 \\ & \text { s.t. } \boldsymbol{s}_i \in\{0,1\}^T, \mathbf{1}^{\top} \boldsymbol{s}_i=1
\end{split}
\end{equation}
where $\boldsymbol{W}^k\in\mathbb{R}^{d\times T}$ represents a matrix consisting of all cluster centers and we assume that all the clusterings have $T$ clusters.
$\boldsymbol{s}_i$ is the assignment vector of the $i$-th example and has only one non-zero element. $\boldsymbol{1}$ signifies a column vector with all elements as $1$.
In the training process, we employ k-means to initialize the cluster centers $\boldsymbol{W}^k$ and optimize $s_i$, and the $\boldsymbol{W}^k$ is held constant to prevent a degenerate solution where all examples converge to a single point, resulting in the objective equaling zero. As a result of $\boldsymbol{W}^k$ being constant, decision boundaries are also established. This is because that the decision boundaries act as perpendicular bisectors of adjacent cluster centers, rendering it impossible to aggregate all examples together by optimizing the given Eqn.~\eqref{eq:loss_cluster}.
Therefore, when $\boldsymbol{z}^k_i$ and $\boldsymbol{W}^k$ in Eqn.~\eqref{eq:loss_cluster} are fixed, $s_i$ can be updated as
\begin{equation}
    s_{ij}\leftarrow
    \begin{cases}
        1, & \text{if} \quad j = \arg\min_t \| \boldsymbol{z}^k_i - \boldsymbol{w}^k_t \|_2\\
        0, & \text{otherwise}
    \end{cases}
\end{equation}
where $s_{ij}$ is the $j$-th element of $\boldsymbol{s}_i$ and $\boldsymbol{w}^k_t$ denotes the centroid of the $t$-th cluster for the $k$-th disentangled representation. 

The training process will cease if the change in cluster labels between two consecutive iterations is less than a specified threshold $\delta$. Formally, the stopping criterion can be described as
\begin{equation}
    1 - \frac{1}{n}\sum_{i,j} s_{ij}^e\cdot s_{ij}^{e-1} < \delta
\end{equation}
where $s_{ij}^{e-1}$ and $s_{ij}^e$ are indicators for whether example $x_i$ is assigned to the $j$-th cluster at the $(e-1)$-th and $e$-th iteration, respectively. We empirically set $\delta=0.0005$ in our experiments.
Thereafter, the cluster assignment objective becomes a constraint in the training process, so that the learned image representations can be more suitable for the clustering purpose, such as k-means tasks. 
To summarize,
the loss function of our method is
\begin{equation}
    \mathcal{L} = \underbrace{ \mathcal{L}_{\text{coarse}} + \mathcal{L}_{\text{fine}} }_{\text{E-step}}+ \underbrace{ \mathcal{L}_{\text{cluster}} }_{\text{M-step}}
\end{equation}
During the optimization, we iteratively perform the E-step and M-step. In E-step, we learn $K$ distinct disentangled representations to uncover multiple aspects of data. In M-step, the disentangled representations are then utilized in the cluster assignment learning process, where each set of representations is grouped into $T$ clusters. The knowledge acquired from cluster assignments serves as a constraint in the subsequent E-step, ensuring the clustering purpose in representation learning. 
\section{Experiments}

\subsection{Experimental Setup}

\begin{table}[t]
   \centering
   \caption{Dataset Statistics.}
   \resizebox{0.45\textwidth}{!}{
       \begin{tabular}{cc cc}
       \toprule
        {Datasets}    &   \# Samples & \# Dimensions & \# Clusters \\
        \midrule
        {ALOI} & 288 & 287 & 2;2 \\
        {Card} & 8,029 & 50,176 & 13;4 \\
        {CMUface} & 640 & 15,360 & 4;20;2;4 \\
        {Fruit}  & 105 & 119,025 & 3;3 \\
        {Fruit360} & 4,856 & 10,000 & 4;4 \\
        {StickFig} & 900 & 400 & 3;3 \\
        {C-MNIST} & 10,000 & 1,568 & 10;10 \\
        \bottomrule
       \end{tabular}
   }
   \vspace{-0.1cm}
   \label{tab:dataset}
\end{table}
\begin{table*}[t]
    \centering
    \caption{Quantitative comparison. The significantly best results with 95\% \%confidence are in bold.}
    \resizebox{\textwidth}{!}{
    \begin{tabular}{cc|cc|cc|cc|c c|cc|cc |cc|cc|cc }
    \toprule
        & & \multicolumn{4}{c|}{\textbf{Single Clustering}} & \multicolumn{14}{c}{\textbf{Multiple Clustering}}\\ \midrule
        \multirow{2}{*}{Datasets} & \multirow{2}{*}{Clusterings} & 
        \multicolumn{2}{c|}{DAC}  &\multicolumn{2}{c|}{DCN}  &
     \multicolumn{2}{c|}{MSC}  & \multicolumn{2}{c|}{MCV} & \multicolumn{2}{c|}{ENRC}  & \multicolumn{2}{c|}{iMClusts}  & \multicolumn{2}{c|}{AugDMC}& \multicolumn{2}{c|}{$\beta$-VAE} & \multicolumn{2}{c}{DDMC}\\

         &  & NMI$\uparrow$ & RI$\uparrow$ &NMI$\uparrow$ & RI$\uparrow$ &NMI$\uparrow$ & RI$\uparrow$ & NMI$\uparrow$ & RI$\uparrow$ & NMI$\uparrow$ & RI$\uparrow$ & NMI$\uparrow$ & RI$\uparrow$ & NMI$\uparrow$ & RI$\uparrow$ & NMI$\uparrow$ & RI$\uparrow$ & NMI$\uparrow$ & RI$\uparrow$ \\ 
        \midrule
        
        \multirow{2}{*}{Fruit} & Color & 0.6579 & 0.7893& 0.6724 & 0.7935& 0.6886 & 0.8051 & 0.6266 & 0.7685 & 0.7103 & 0.8511 & 0.7351 & 0.8632 & \underline{0.8517} & \underline{0.9108} & 0.8329 & 0.8611 &
        \textcolor{blue}{\textbf{0.8973}} & \textcolor{blue}{\textbf{0.9383}} \\ 
        ~ & Species & 0.2154 & 0.6206& 0.2058 & 0.6127& 0.1627 & 0.6045 & 0.2733 & 0.6597 & 0.3187 & 0.6536 & 0.3029 & 0.6743 & \underline{0.3546} & \underline{0.7399} & 0.3287 & 0.6562 & \textcolor{blue}{\textbf{0.3764}} & \textcolor{blue}{\textbf{0.7621}} \\ \midrule
        
        \multirow{2}{*}{Fruit360} & Color & 0.1967 & 0.5568& 0.2197 & 0.5899 & 0.2544 & 0.6054& 0.3776 & 0.6791 & 0.4264 & 0.6868 & 0.4097 & 0.6841 & \underline{0.4594} & \underline{0.7392} & 0.4354 & 0.7043 & \textcolor{blue}{\textbf{0.4981}} & \textcolor{blue}{\textbf{0.7472}} \\ 
        ~ & Species & 0.1533 & 0.5439& 0.1685 & 0.5583& 0.2184 & 0.5805 & 0.2985 & 0.6176 & 0.4142 & 0.6984 & 0.3861 & 0.6732 & \underline{0.5139} & \underline{0.7430} & 0.4289 & 0.6982 &  \textcolor{blue}{\textbf{0.5292}} & \textcolor{blue}{\textbf{0.7703}} \\ \midrule

        \multirow{2}{*}{Card} & Order & 0.0532 & 0.6392& 0.0612 & 0.6893& 0.0807 & 0.7805 & 0.0792 & 0.7128 & 0.1225 & 0.7313 & 0.1144 & 0.7658 & \underline{0.1440} & \underline{0.8267} & 0.1205 & 0.7329 & \textcolor{blue}{\textbf{0.1563}} & \textcolor{blue}{\textbf{0.8326}} \\ 
        ~ & Suits & 0.0269 & 0.3350 & 0.0416 & 0.3604& 0.0497 & 0.3587 & 0.0430 & 0.3638 & 0.0676 & 0.3801 & 0.0716 & 0.3715 & \underline{0.0873} & 0.4228 & 0.0728 & \underline{0.5536} & \textcolor{blue}{\textbf{0.0933}} & \textcolor{blue}{\textbf{0.6469}} \\ \midrule

        \multirow{2}{*}{StickFig} & Upper & 0.3781 & 0.5964& 0.3873 & 0.6390& 0.6293 & 0.7293 & 0.5387 & 0.6896 & 1.0000 & 1.0000 & 1.0000 & 1.0000 & 1.0000 & 1.0000 & 0.8226 & 0.9033 & 1.0000 & 1.0000 \\ 
        ~ & Lower & 0.3695 & 0.6009 & 0.3651 & 0.6338& 0.6431 & 0.7149 & 0.5160 & 0.6524 & 1.0000 & 1.0000 & 1.0000 & 1.0000 & 1.0000 & 1.0000 & 0.8261 & 0.9084 & 1.0000 & 1.0000 \\ \midrule

        \multirow{2}{*}{ALOI} & Shape & 0.2172 & 0.5153& 0.3839 & 0.7391& 0.2968 & 0.5199 & 0.7359 & 0.8261 & 0.9732 & 0.9861 & 0.9963 & 0.9989 & 1.0000 & 1.0000 & 0.9706 & 0.9887 & 1.0000 & 1.0000 \\ 
        ~ & Color & 0.1520 & 0.3893& 0.2083 & 0.5820& 0.1563 & 0.3428 & 0.6982 & 0.7439 & 0.9833 & 0.9892 & 1.0000 & 1.0000 & 1.0000 & 1.0000 & 0.9754 & 0.9892& 1.0000 & 1.0000 \\ \midrule
        \multirow{4}{*}{CMUface} & Emotion & 0.0612 & 0.5157& 0.0811 & 0.5107& 0.1284 & \underline{0.6736} & 0.1433 & 0.5268 & \underline{0.1592} & 0.6630 & 0.0422 & 0.5932 & 0.0161 & 0.5367 & 0.1549 & 0.6580 &  \textcolor{blue}{\textbf{0.1726}} & \textcolor{blue}{\textbf{0.7593}} \\ 
        ~ & Glass & 0.0439 & 0.4692& 0.0535 & 0.4791& 0.1420 & 0.5745 & 0.1201 & 0.4905 & 0.1493 & 0.6209 & \underline{0.1929} & 0.5627 & 0.1039 & 0.5361 & 0.1897 & \underline{0.6225}& \textcolor{blue}{\textbf{0.2261}} & \textcolor{blue}{\textbf{0.7663}} \\ 
        ~ & Identity & 0.4196 & 0.7653& 0.4912 & 0.7932& 0.3892 & 0.7326 & 0.4637 & 0.6247 & 0.5607 & 0.7635 & 0.5109 & 0.8260 & \underline{0.5876} & \underline{0.8334} & 0.4535& 0.6873 &\textcolor{blue}{\textbf{0.6360}} & \textcolor{blue}{\textbf{0.8907}} \\ 
        ~ & Pose & 0.2184 & 0.5524& 0.2306 & 0.5596& 0.3687 & 0.6322 & 0.3254 & 0.6028 & 0.2290 & 0.5029 & \underline{0.4437} & 0.6114 & 0.1320 & 0.5517 & 0.3882 & \underline{0.6831} & \textcolor{blue}{\textbf{0.4526}} & \textcolor{blue}{\textbf{0.7904}} \\ \midrule

        \multirow{2}{*}{C-MNIST} & Left & 0.0857 & 0.5235& 0.1038 & 0.5283& 0.0167 & 0.6273 & 0.1326 & 0.5603 & 0.7263 & 0.7882 & 0.7736 & 0.8250 & \underline{0.9364} & \underline{0.9569} & 0.8085 & 0.9105 & \textcolor{blue}{\textbf{1.0000}} & \textcolor{blue}{\textbf{1.0000}} \\ 
        ~ & Right & 0.0828 & 0.5185& 0.0985 & 0.5534& 0.0542 & 0.6003 & 0.1103 & 0.5938 & 0.7277 & 0.7926 & 0.7698 & 0.8119 & \underline{0.9277} & \underline{0.9208} & 0.8123 & 0.8907 & \textcolor{blue}{\textbf{1.0000}} & \textcolor{blue}{\textbf{1.0000}} \\ 
        \bottomrule
    \end{tabular}
    }
    \label{tab:clustering}
    \vspace{-0.3cm}
\end{table*}
\begin{table*}
    \centering
    \caption{Components ablation. The significantly best results with 95\% confidence are in bold.}
    \resizebox{\textwidth}{!}{
    \begin{tabular}{cc|cc|c c|cc|cc| cc|cc}
    \toprule
        \multirow{2}{*}{Datasets} & \multirow{2}{*}{Clusterings} & \multicolumn{2}{c|}
    {\sysname$_{wo\text{mix}}$} & \multicolumn{2}{c|}{\sysname$_{{w}=0.5}$} & \multicolumn{2}{c|}{DDMC$_{wo\text{CD}}$} & \multicolumn{2}{c|}{DDMC$_{wo\text{CA}}$ }& \multicolumn{2}{c|}{DDMC$_{wo\text{CD\&CA}}$} & \multicolumn{2}{c}{DDMC} \\ 
        ~ & ~ & NMI$\uparrow$ & RI$\uparrow$ & NMI$\uparrow$ & RI$\uparrow$ & NMI$\uparrow$ & RI$\uparrow$ & NMI$\uparrow$ & RI$\uparrow$ & NMI$\uparrow$ & RI$\uparrow$ & NMI$\uparrow$ & RI$\uparrow$ \\ \midrule
        \multirow{2}{*}{Fruit} & Color & 0.8854 & \underline{0.9013} & \underline{0.8912} & 0.9010 & 0.8796 & 0.8993 & 0.8869 & 0.8981 & 0.8621 & 0.8824 & \textcolor{blue}{\textbf{0.8973}} & \textcolor{blue}{\textbf{0.9383}} \\ 
        ~ & Species & 0.3610 & 0.7389 & \underline{0.3695} & \underline{0.7429} & 0.3592 & 0.7268 & 0.3557 & 0.7369 & 0.3428 & 0.7007 & \textcolor{blue}{\textbf{0.3764}} & \textcolor{blue}{\textbf{0.7621}} \\ 
        \midrule
        \multirow{2}{*}{Fruit360} & Color & 0.4887 & 0.7315 & \underline{0.4923} & \underline{0.7399} & 0.4754 & 0.7209 & 0.4862 & 0.7277 & 0.4633 & 0.7203 & \textcolor{blue}{\textbf{0.4981}} & \textcolor{blue}{\textbf{0.7472}} \\ 
        ~ & Species & 0.5218 & 0.7494 & \underline{0.5254} & \underline{0.7584} & 0.5037 & 0.7353 & 0.5138 & 0.7508 & 0.4909 & 0.7285 & \textcolor{blue}{\textbf{0.5292}} & \textcolor{blue}{\textbf{0.7703}} \\ 
        \midrule
        \multirow{2}{*}{Card} & Order & 0.1482 & 0.7927 & \underline{0.1527} & 0.8217 & 0.1308 & 0.7671 & 0.1513 & \underline{0.8285} & 0.1253 & 0.7523 & \textcolor{blue}{\textbf{0.1563}} & \textcolor{blue}{\textbf{0.8326}} \\ 
        ~ & Suits & 0.0899 & 0.6326 & \underline{0.0918} & \underline{0.6382} & 0.0779 & 0.5982 & 0.0896 & 0.6297 & 0.0730 & 0.5508 & \textcolor{blue}{\textbf{0.0933}} & \textcolor{blue}{\textbf{0.6469}} \\ 
        \midrule
        \multirow{2}{*}{StickFig} & Upper & 1.0000 & 1.0000 & 1.0000 & 1.0000 & 0.8568 & 0.9226 & 1.0000 & 1.0000 & 0.8364 & 0.9186 & 1.0000 & 1.0000 \\ 
        ~ & Lower & 1.0000 & 1.0000 & 1.0000 & 1.0000 & 0.8490 & 0.9094 & 1.0000 & 1.0000 & 0.8436 & 0.8922 & 1.0000 & 1.0000 \\ 
        \midrule
        \multirow{2}{*}{ALOI} & Shape & 0.9912 & 0.9956 & 1.0000 & 1.0000 & 0.9891 & 0.9921 & 1.0000 & 1.0000 & 0.9885 & 0.9931 & 1.0000 & 1.0000 \\ 
        ~ & Color & 0.9927 & 0.9952 & 1.0000 & 1.0000 & 0.9853 & 0.9956 & 1.0000 & 1.0000 & 0.9798 & 0.9889 & 1.0000 & 1.0000 \\ 
        \midrule
        \multirow{4}{*}{CMUface} & Emotion & 0.1698 & 0.7452 & \underline{0.1709} & \underline{0.7483} & 0.1707 & 0.7382 & 0.1698 & 0.7436 & 0.1674 & 0.7045 & \textcolor{blue}{\textbf{0.1726}} & \textcolor{blue}{\textbf{0.7593}} \\ 
        ~ & Glass & 0.2205 & 0.7598 & 0.2249 & \underline{0.7621} & 0.2253 & 0.7568 & 0.2243 & 0.7514 & \underline{0.2262} & 0.7208 & \textcolor{blue}{\textbf{0.2261}} & \textcolor{blue}{\textbf{0.7663}} \\ 
        ~ & Identity & 0.5806 & 0.8570 & 0.6014 & 0.8792 & 0.5294 & 0.7932 & \underline{0.6317} & \underline{0.8823} & 0.4736 & 0.7617 & \textcolor{blue}{\textbf{0.6360}} & \textcolor{blue}{\textbf{0.8907}} \\ 
        ~ & Pose & 0.4492 & 0.7723 & \underline{0.4510} & \underline{0.7855} & 0.4381 & 0.7360 & 0.4427 & 0.7806 & 0.4022 & 0.7183 & \textcolor{blue}{\textbf{0.4526}} & \textcolor{blue}{\textbf{0.7904}} \\ 
        \midrule
        \multirow{2}{*}{C-MNIST} & Left & 1.0000 & 1.0000 & 1.0000 & 1.0000 & 0.8562 & 0.9253 & 1.0000 & 1.0000 & 0.8259 & 0.9020 & 1.0000 & 1.0000 \\ 
        ~ & Right & 1.0000 & 1.0000 & 1.0000 & 1.0000 & 0.8495 & 0.9190 & 1.0000 & 1.0000 & 0.8305 & 0.8994 & 1.0000 & 1.0000 \\ \bottomrule
    \end{tabular}
    }
    \label{tab:ablation}
    \vspace{-0.3cm}
\end{table*}

\subsubsection{Datasets}

To demonstrate our proposed method, we evaluate \sysname ~on seven benchmark image datasets in multiple clustering which are summarized in Table~\ref{tab:dataset}. It is noteworthy that the notation for \#clusters implies that there are several clusters under each clustering.
\begin{itemize}
\item \textbf{ALOI}~\cite{geusebroek2005amsterdam} (Amsterdam Library of Object Images) contains images of 1000 common objects. Following setting in~\cite{ren2022diversified}, we sample 288 images of four objects with two clusterings: color (yellow and red) and shape (circle and cylinder).
\item \textbf{Card}\footnote{https://www.kaggle.com/datasets/gpiosenka/cards-image-datasetclassification} is a dataset of playing card images, which consists of 8,029 images with two clusterings, i.e., rank (Ace, King, Queen, etc.) and suits (clubs, diamonds, hearts, spades).
\item \textbf{CMUface}~\cite{gunnemann2014smvc} contains 640 $32 \times 30$ gray images, which can be grouped according to pose (left, right, straight and up), identity (20 individuals), glass (with or without glass), and emotions (angry, happy, neutral and sad).
\item \textbf{Fruit}~\cite{hu2017finding} consists of 105 images and has two clusterings, i.e., species (apples, bananas, and grapes) and color (green, red, and yellow).
\item \textbf{Fruit360}\footnote{https://www.kaggle.com/moltean/fruits} contains 4,856 images and has two clusterings as well, i.e., species (apples, bananas, cherries, and grapes) and color (red, green, yellow, and maroon). 
\item \textbf{StickFig}~\cite{gunnemann2014smvc} has 900 $20 \times 20$ images and two clusterings, i.e., upper body and lower body. Each clustering has three clusters according to the body postures.
\item \textbf{MNIST}~\cite{lecun1998gradient} is a well known dataset. Here, we extend it through concatenating two digits side by side, leading to a total of 100 possible combinations, named C-MNIST. As a result, the  dataset can be seen as containing two clusterings, i.e., left and right. Each of the clusterings has 10 clusters.

\end{itemize}

\subsubsection{Baselines}
We compare \sysname{} against eight state-of-the-art methods including two single clustering methods and six multiple clustering methods: \textbf{DAC}~\cite{chang2017deep} is a deep single clustering method that adopts a pairwise classification framework to learn image representations for clustering; \textbf{DCN}~\cite{yang2017towards} learns feature representations through a clustering constraint, making the representations more suitable for single clustering tasks; \textbf{MSC}~\cite{hu2017finding} is a traditional multiple clustering method that uses hand-crafted features as shown by \#Dimensions in Table~\ref{tab:dataset}; \textbf{MCV}~\cite{guerin2018improving} leverages multiple pre-trained feature extractors to represent different ``views’’ of the same data and employs a multi-input neural network to enhance clustering outcomes; \textbf{ENRC}~\cite{miklautz2020deep} is a deep multiple clustering method that integrates auto-encoder and clustering objective to generate different clusterings; \textbf{iMClusts}~\cite{ren2022diversified} makes use of the expressive representational power of deep autoencoders and multi-head attention to accomplish multiple clusterings; \textbf{AugDMC}~\cite{DBLP:conf/inns-dlia/YaoLRH23} leverages augmentations to learn different image representations to achieve multiple clustering; and \textbf{$\beta$-VAE}~\cite{higgins2017beta} is a disentangled representation method based on VAE to learn distinct representations in an unsupervised manner, whose disentangled representations are directly used for multiple clustering. As our method, $\beta$-VAE also uses k-means~\cite{lloyd1982least} as the clustering method.

\subsubsection{Implementation Setup}
In implementation, ResNet-18~\cite{he2016deep} is adopted as the encoder and decoder. 
Several data augmentation methods are utilized in the experiments, such as ``RandomRotation'', ``RandomHorizontalFlip'', ``RandomCrop'', ``ColorJitter'', and so on. We employ Adam and set momentum as $0.9$ to train the model for $1000$ epochs. All hyperparameters are searched according to the loss score of \sysname{}, where the learning rate is searched in $\{0.1,0.05,0.01,0.005,0.001,0.0005\}$, weight decay is in $\{0.0005, 0.0001, 0.00005, 0.00001, 0\}$, and temperature $\tau$ is in $\{0.8,0.85,0.9,0.95,1.0\}$. We perform k-means~\cite{lloyd1982least} 10 times due to its randomness and evaluate the average clustering performance using two quantitative metrics, that is, Normalized Mutual Information (NMI)~\cite{white2004performance} and Rand index (RI)~\cite{rand1971objective}. These measures range in $[0, 1]$, and higher scores imply more accurate clustering results.
The experiments are conducted with a single GPU NVIDIA GeForce RTX 2080 Ti.

\subsection{Performance Comparison}
Table~\ref{tab:clustering} presents the average clustering results, with the significantly best results highlighted in bold, and the second-best results underlined. \sysname{} consistently outperforms, indicating its superiority, which is demonstrated from the following aspects. \textit{First}, comparing single to multiple clustering, we can observe that the single clustering methods, namely DAC and DCN, typically underperform in comparison to their multiple clustering counterparts. This is an expected outcome as single clustering methods struggle to discern distinct clusters within datasets. Furthermore, DCN often outperforms DAC. This can be attributed to the fact that DCN leverages an optimized cluster objective function, which aids DCN in identifying better data grouping.
\textit{Second}, comparing between deep and shallow multiple clustering, the deep methods, namely MCV, ENRC, iMClusts, AugDMC, $\beta$-VAE and \sysname{}, yield superior performance than MSC, especially on large-scale datasets like C-MNIST and Fruit360. These results not only demonstrate the effectiveness of deep learning models in multiple clustering but also underline the potential of deep neural networks in uncovering underlying cluster structures. 
\textit{Third}, disentangled-based methods, $\beta$-VAE and \sysname{}, generally yield superior results. Despite $\beta$-VAE not being tailored for multiple clustering, it delivers strong performance due to its capability to disentangle latent factors. This underscores the efficacy of disentangled representation learning in multiple clustering. Furthermore, \sysname{} outperforms $\beta$-VAE, lending credibility to the effectiveness of our \sysname{}. 
\textit{Finally}, both AugDMC and \sysname{} are methods that employ augmentation to learn diverse representations. These methods outperform other techniques in most cases, suggesting that data augmentation can preserve distinct aspects of the data and subsequently aid in discovering diverse features for multiple clustering. 
More importantly, \sysname{} surpasses AugDMC in most cases, largely owing to its more sophisticated and robust design that incorporates fine-grained disentangled learning and cluster assignment. Notably, AugDMC does not perform well on the CMUface dataset. This is likely because that the clusters in this data are related to detailed features such as emotion and glasses, which are challenging to capture with augmentation methods. Despite this, the proposed \sysname{} method continues to deliver the best results, thanks to its fine-grained disentangled representation and cluster assignment components.

\begin{figure}[t]
    \centering
    \subfigure[\sysname{}$_{wo\text{mix}}$]{
        \includegraphics[width=0.2\textwidth]{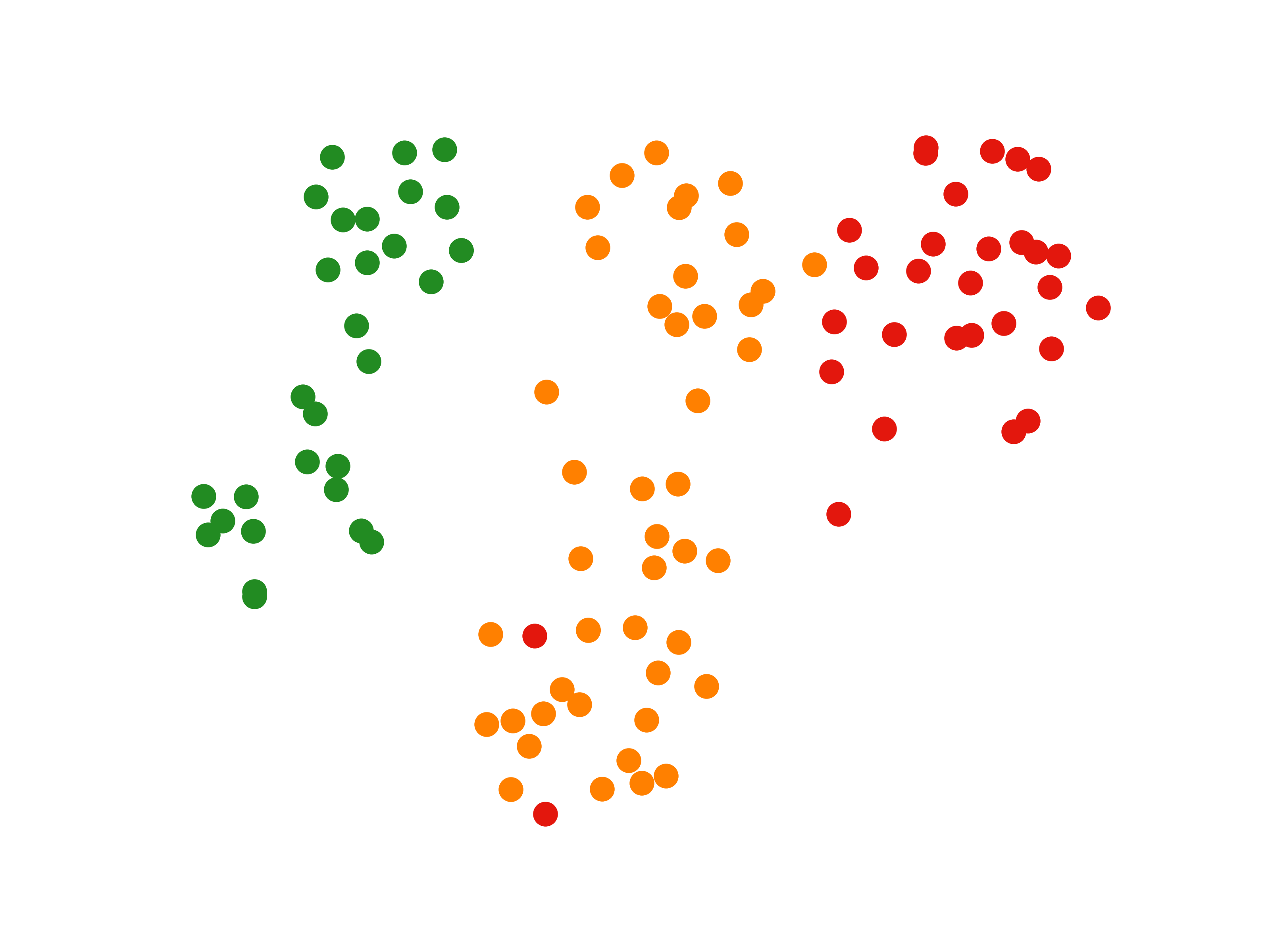}
    }
    \subfigure[\sysname{}$_{w=0.5}$]{
        \includegraphics[width=0.2\textwidth]{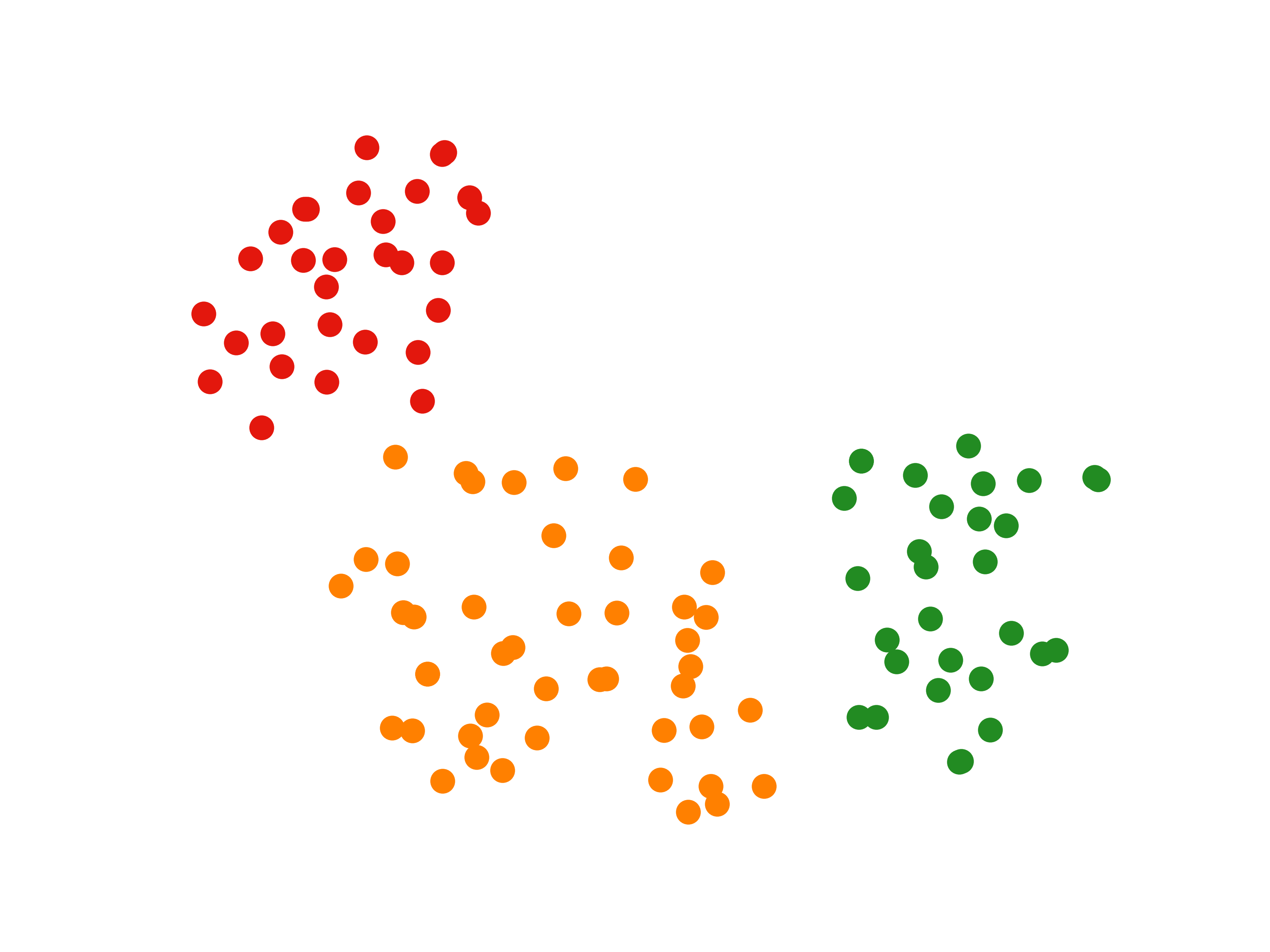}
    }
    \subfigure[\sysname{}$_{wo\text{CD}}$]{
        \includegraphics[width=0.2\textwidth]{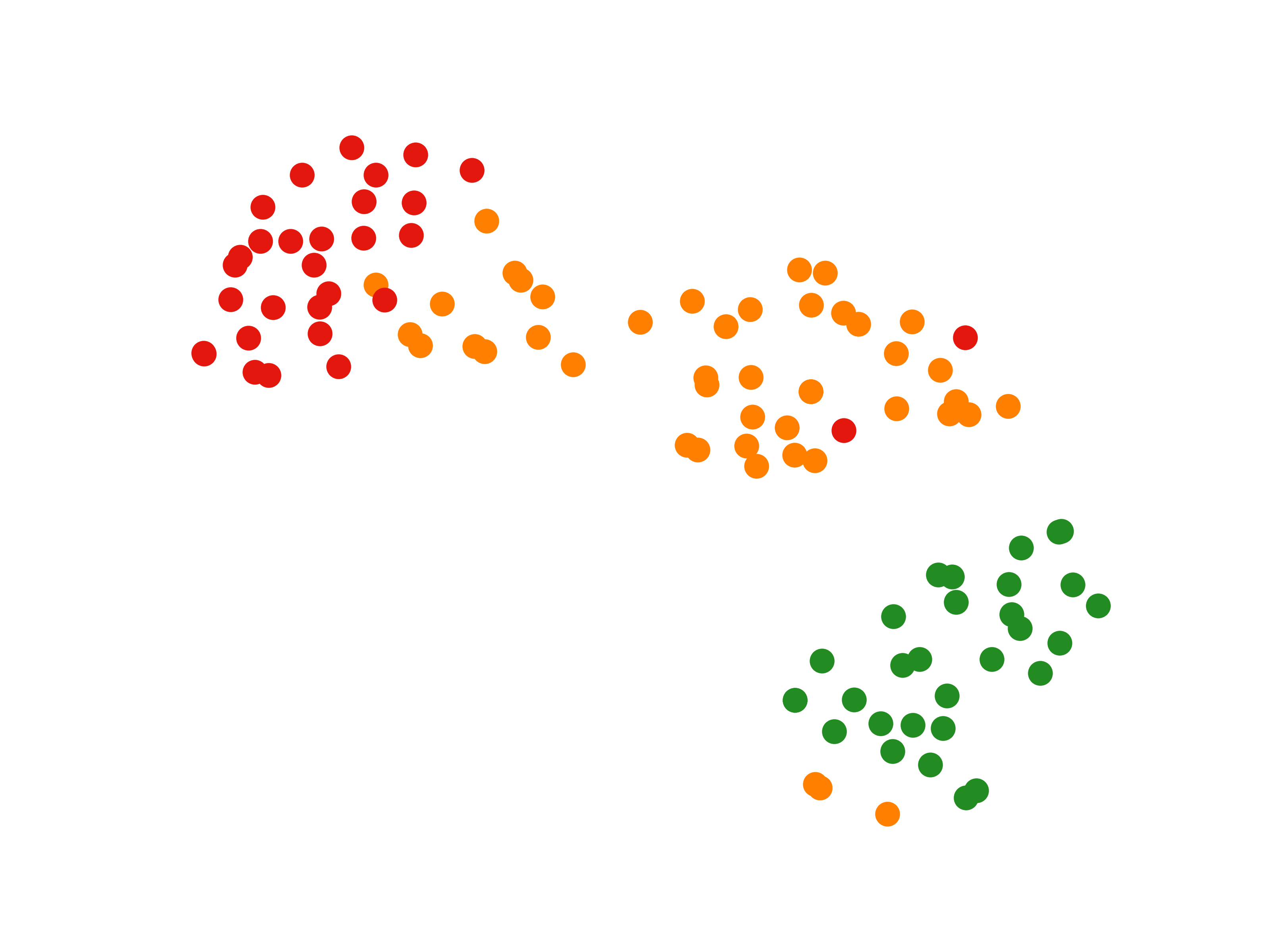}
    }
    \subfigure[\sysname{}$_{wo\text{CA}}$]{
        \includegraphics[width=0.2\textwidth]{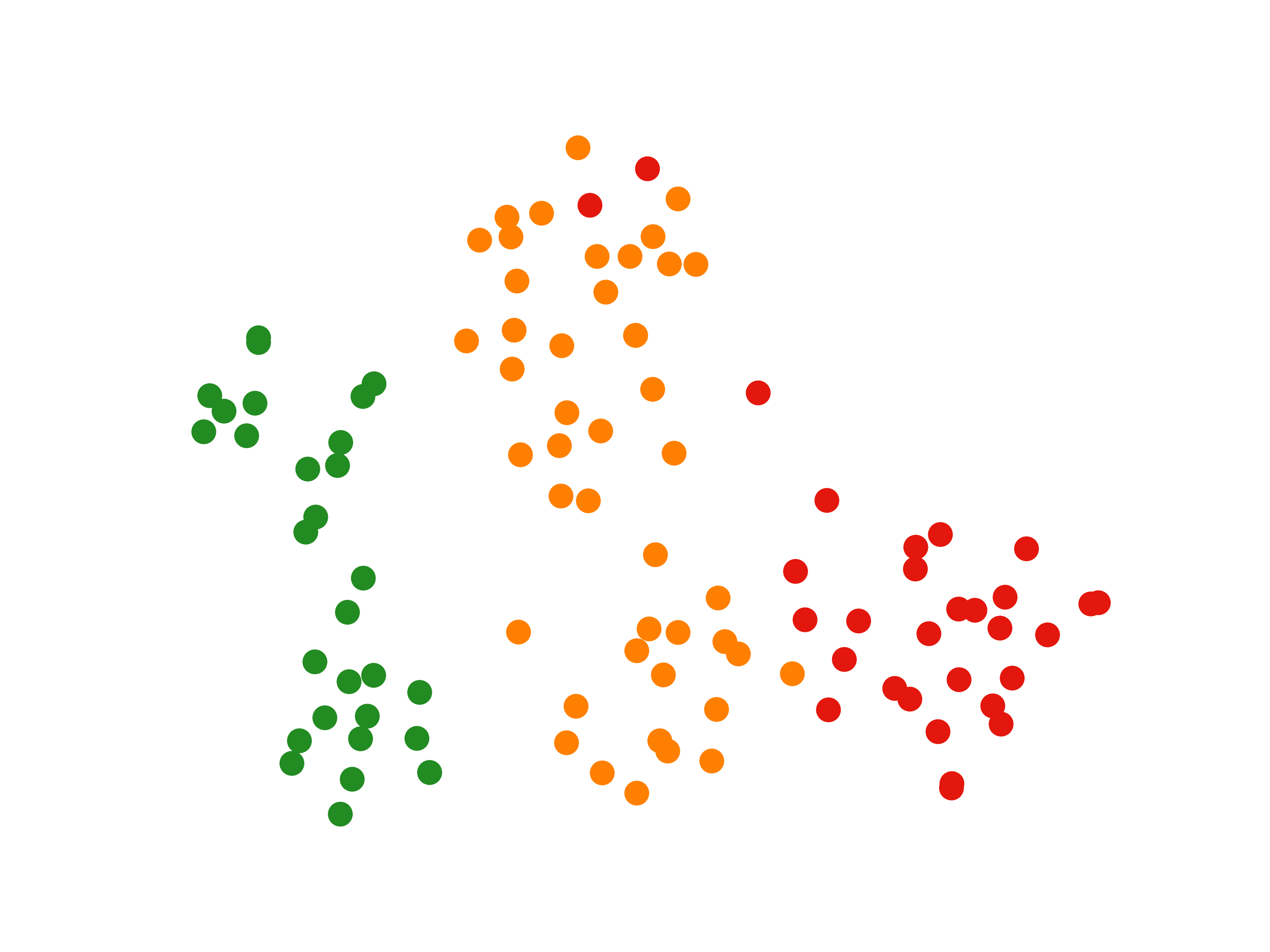}
    }
    \subfigure[\sysname{}$_{wo\text{CD\&CA}}$]{
        \includegraphics[width=0.2\textwidth]{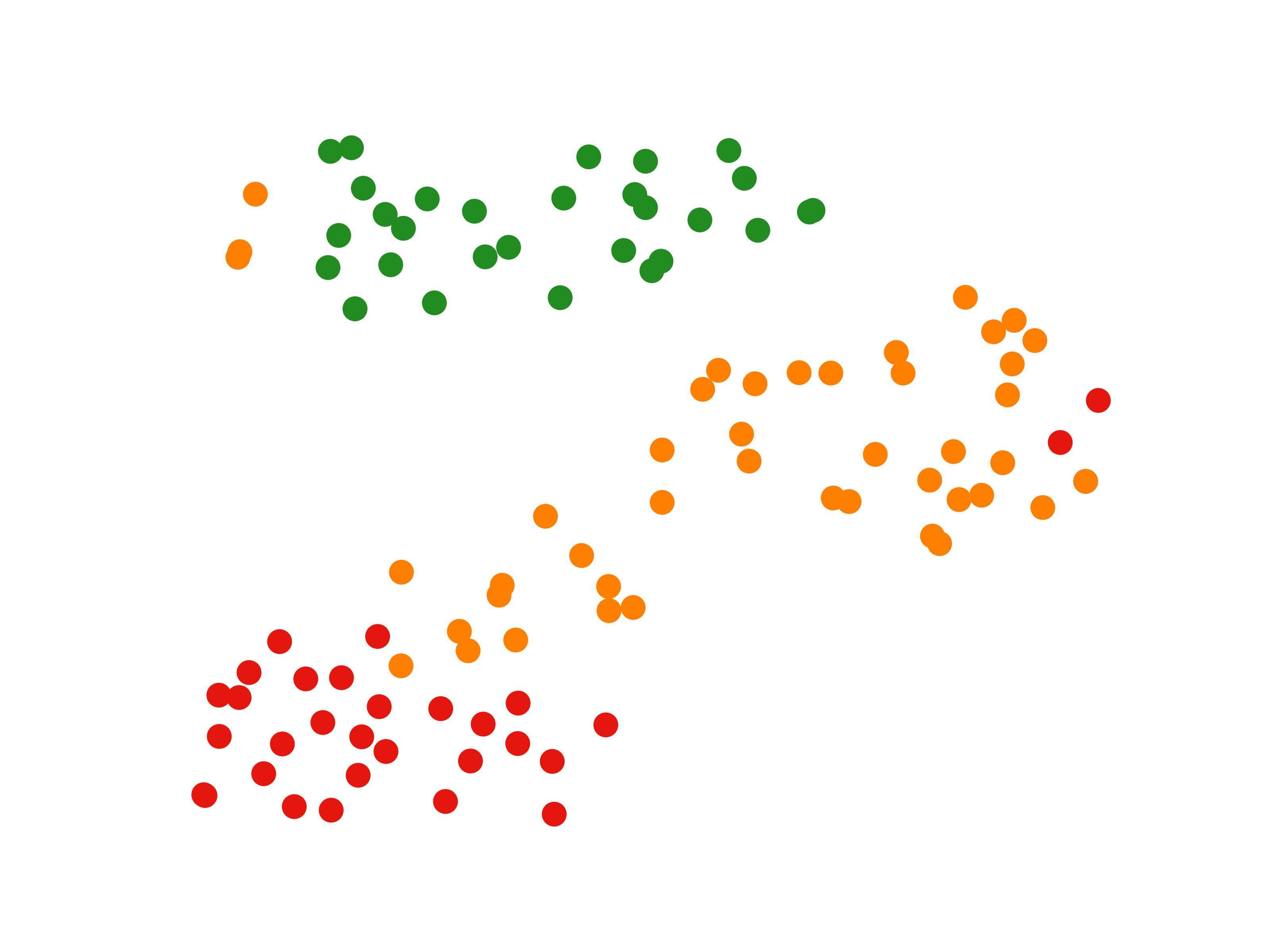}
    }
    \subfigure[\sysname{}]{
        \includegraphics[width=0.2\textwidth]{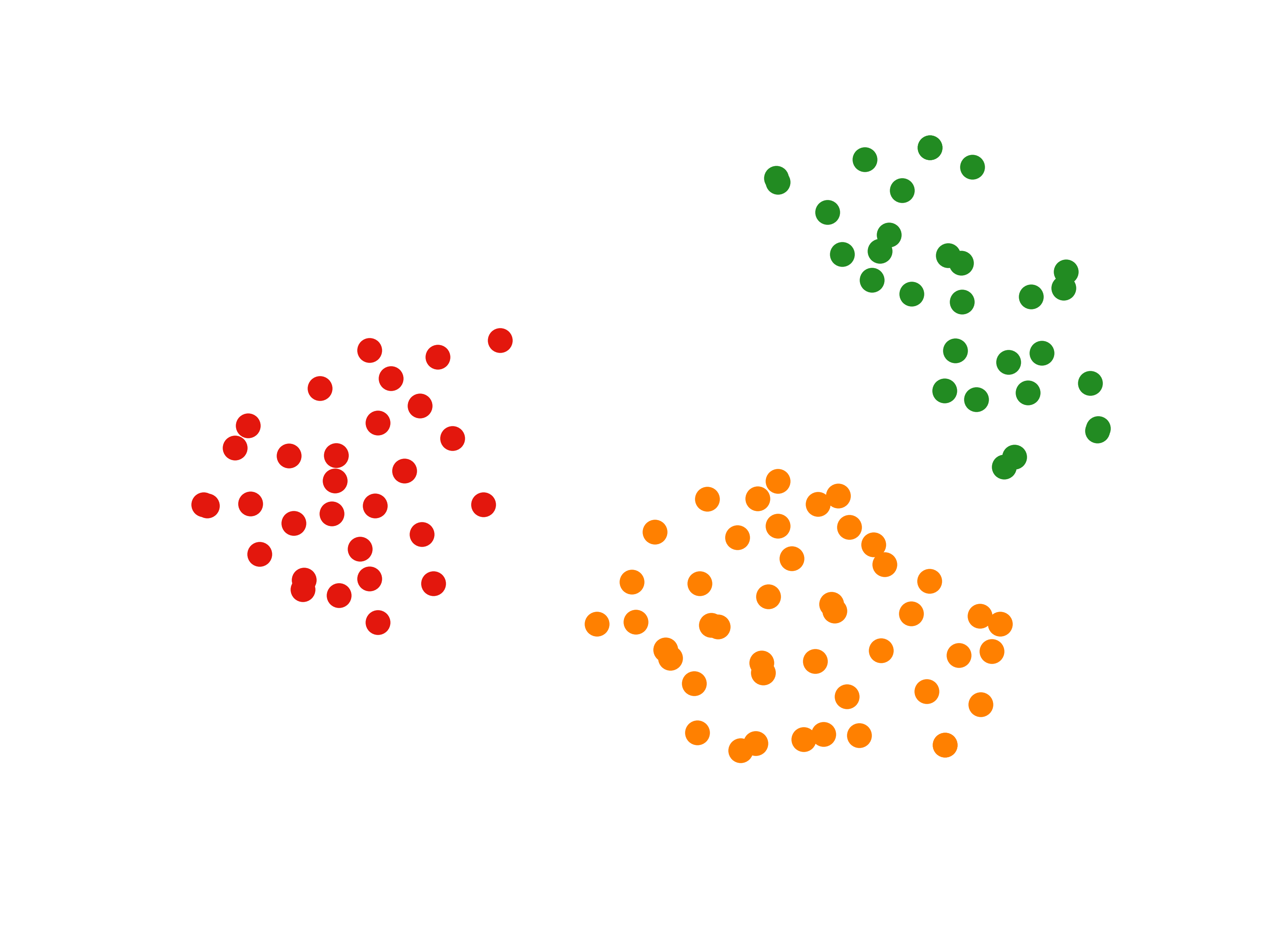}
    }
    
    \caption{Visualization of \sysname{} and its variants color representations on the Fruit dataset. 
    }
    \label{fig:vis1}
\vspace{-0.7cm}
\end{figure}

\subsection{Ablation Study}
We also assess the contribution of different components within \sysname{}.
Specifically, to verify the effect of mixing in the coarse-grained disentangled learning, we remove the mixing process and set ${w}=0.5$ to show the effect of the trainable parameter ${w}$. Consequently, we can obtain two related variants referred as 
\sysname$_{wo\text{mix}}$ and \sysname$_{{w}=0.5}$, respectively.
Moreover, we also generate three additional variants by removing the coarse-grained disentangled representation, the cluster assignment component, or both as \sysname$_{wo\text{CD}}$, \sysname$_{wo\text{CA}}$, and \sysname$_{wo\text{CD\&CA}}$, respectively. The results are illustrated in Table~\ref{tab:ablation}. \sysname{} demonstrates superior performance in all cases, showing the effectiveness of both the coarse-grained disentangled representation and cluster assignment. Specifically, 
\sysname{}$_{w=0.5}$ outperforms \sysname{}$_{wo\text{mix}}$, indicating the efficacy of the mixing process within the coarse-grained disentanglement. \sysname{}$_{wo\text{mix}}$ differs \sysname{}$_{wo\text{CD}}$ in the augmentation process. \sysname{}$_{wo\text{mix}}$ achieves better results than \sysname{}$_{wo\text{CD}}$ in most cases, implying that augmentation is advantageous in learning diverse representations. However, \sysname{}$_{wo\text{CD}}$ outperforms both \sysname{}$_{wo\text{mix}}$ and \sysname{}$_{w=0.5}$ on the emotion and glass clustering on CMUface. This can be attributed to the difficulty in capturing intricate information through augmentation. Besides, \sysname{} outperforms both \sysname{}$_{wo\text{CD}}$ and \sysname{}$_{wo\text{CA}}$, confirming the efficacy of coarse-grained disentanglement and cluster assignment, respectively. This finding is echoed in the comparison between \sysname{} and \sysname{}$_{wo\text{CD\&CA}}$. 

\subsection{Visualization}
We further conduct the visual comparsion for \sysname{} and its variants on the Fruit dataset using t-SNE~\cite{van2008visualizing}. As shown in Fig.~\ref{fig:vis1}, We use red, yellow, and green points to denote images with red, yellow, and green labels, respectively. Consistent with the results observed in the ablation study, compared with other variants, \sysname{} can establish clearer boundaries between data points of different categories and form a more compact distribution within the same category. These results further demonstrate the effectiveness of both cluster assignment and coarse-grained disentanglement representation.

\subsection{Clustering Analysis}
We delve deeper into the analysis of the clustering outcomes of \sysname{} against various ground truth label sets. These results are derived by applying different disentangled representations to produce all clustering outputs. The findings, as depicted in Table~\ref{tab:cluster_analysis} and Table~\ref{tab:cluster_analysis2}, reveal that the top-performing outcomes exhibit a clear diagonal structure. This demonstrates that the disentangled representations are capable of discerning unique clusterings within the same dataset.

\begin{table}
    \centering
    \caption{Clustering analysis on six datasets. The best results are in bold.}
    \resizebox{0.45\textwidth}{!}{
    \begin{tabular}{c|c|cc|cc}
    \toprule
        \multirow{2}{*}{Datasets} & \multirow{2}{*}{Clusterings} & \multicolumn{2}{c|}{ $C^1$}  & \multicolumn{2}{c}{$C^2$}\\ 
        ~ & ~ & NMI & RI & NMI & RI \\ \midrule
        \multirow{2}{*}{Fruit} & Color & \textcolor{blue}{\textbf{0.8973}} & \textcolor{blue}{\textbf{0.9383}} & 0.8256 & 0.8842 \\ 
        ~ & Species & 0.3528 & 0.7026 & \textcolor{blue}{\textbf{0.3764}} & \textcolor{blue}{\textbf{0.7621}} \\ \midrule
        \multirow{2}{*}{Fruit360} & Color & \textcolor{blue}{\textbf{0.4981}} & \textcolor{blue}{\textbf{0.7472}} & 0.3629 & 0.6860 \\ 
        ~ & Species & 0.3967 & 0.7234 & \textcolor{blue}{\textbf{0.5292}} & \textcolor{blue}{\textbf{0.7703}} \\ \midrule
        \multirow{2}{*}{Card} & Order & \textcolor{blue}{\textbf{0.1563}} & \textcolor{blue}{\textbf{0.8326}} & 0.0961 & 0.5395 \\ 
        ~ & Suits & 0.0526 & 0.5506 & \textcolor{blue}{\textbf{0.0933}} & \textcolor{blue}{\textbf{0.6469}} \\ \midrule
        \multirow{2}{*}{StickFig} & Upper & \textcolor{blue}{\textbf{1.0000}} & \textcolor{blue}{\textbf{1.0000}} & 0.1308 & 0.7009 \\ 
        ~ & Lower & 0.1391 & 0.6848 & \textcolor{blue}{\textbf{1.0000}} & \textcolor{blue}{\textbf{1.0000}} \\ \midrule
        \multirow{2}{*}{ALOI}& Shape & \textcolor{blue}{\textbf{1.0000}} & \textcolor{blue}{\textbf{1.0000}} & 0.1864 & 0.6173 \\ 
        ~ & Color & 0.1755 & 0.6292 & \textcolor{blue}{\textbf{1.0000}} & \textcolor{blue}{\textbf{1.0000}} \\ \midrule
        \multirow{2}{*}{C-MNIST} & Left & \textcolor{blue}{\textbf{1.0000}} & \textcolor{blue}{\textbf{1.0000}} & 0.0798 & 0.6105 \\ 
        ~ & Right & 0.0826 & 0.6568 & \textcolor{blue}{\textbf{1.0000}} & \textcolor{blue}{\textbf{1.0000}} \\ \hline
    \end{tabular}}
    \label{tab:cluster_analysis}
\end{table}
\begin{table}[!t]
    \centering
    \caption{Clustering analysis on CMUface. The best results are in bold.}
       \resizebox{0.45\textwidth}{!}{
    \begin{tabular}{c|c|cccc}
    \toprule
        Clusterings & Metrics & $C^1$ & $C^2$ & $C^3$ & $C^4$ \\\midrule
        \multirow{2}{*}{Emotion} & NMI & \textcolor{blue}{\textbf{0.1726}} & 0.0811 & 0.0006 & 0.0064 \\ 
        ~ & RI & \textcolor{blue}{\textbf{0.7593}} & 0.5538 & 0.5371 & 0.5428 \\ \midrule
        \multirow{2}{*}{Glass} & NMI & 0.0541 & \textcolor{blue}{\textbf{0.2261}} & 0.0039 & 0.0541 \\ 
        ~ & RI & 0.5355 & \textcolor{blue}{\textbf{0.7663}} & 0.5005 & 0.5538 \\ \midrule
        \multirow{2}{*}{Identity} & NMI & 0.3621 & 0.1726 & \textcolor{blue}{\textbf{0.6360}} & 0.4963 \\ 
        ~ & RI & 0.7936 & 0.6760 & \textcolor{blue}{\textbf{0.8907}} & 0.8869 \\ \midrule
        \multirow{2}{*}{Pose} & NMI & 0.0183 & 0.0702 & 0.0468 & \textcolor{blue}{\textbf{0.4526}} \\ 
        ~ & RI & 0.6036 & 0.5878 & 0.6282 & \textcolor{blue}{\textbf{0.7904}} \\ \hline
    \end{tabular}
    }
    \label{tab:cluster_analysis2}
\end{table}

\subsection{Parameter Analysis}

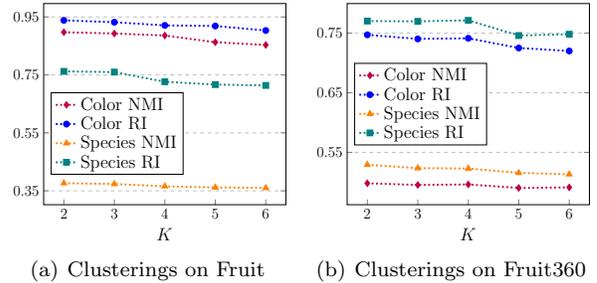
\begin{figure}[t!]
\centering
\subfigure[Clusterings on Fruit]{
            \begin{tikzpicture}[font=\Large, scale=0.47]
                \begin{axis}[
                    legend cell align={left},
                    legend style={nodes={scale=1.0, transform shape}, at={(0.03,0.15)},anchor=south west},
                    xlabel={$K$},
                    xtick pos=left,
                    tick label style={font=\large},
                    ylabel style={font=\large},
                    xtick={0.8, 0.85, 0.9, 0.95, 1.0},
                    xticklabels={$2$,$3$,$4$,$5$,$6$},
                    ytick={0.35, 0.55,0.75,0.95},
                    yticklabels={ $0.35$,$0.55$,$0.75$,$0.95$},
                    ymajorgrids=true,
                    grid style=dashed
                ]
                \addplot[
                    color=purple,
                    dotted,
                    mark options={solid},
                    mark=diamond*,
                    line width=1.5pt,
                    mark size=2pt
                    ]
                    coordinates {
                    (0.8, 0.8973)
                    (0.85, 0.8929)
                    (0.9, 0.8859)
                    (0.95, 0.8627)
                    (1.0, 0.8533)
                    };
                    \addlegendentry{Color NMI}
                \addplot[
                    color=blue,
                    dotted,
                    mark options={solid},
                    mark=*,
                    line width=1.5pt,
                    mark size=2pt
                    ]
                    coordinates {
                    (0.8, 0.9383)
                    (0.85, 0.9318)
                    (0.9, 0.9208)
                    (0.95, 0.9187)
                    (1.0, 0.9033)
                    };
                    \addlegendentry{Color RI}
                \addplot[
                    color=orange,
                    dotted,
                    mark options={solid},
                    mark=triangle*,
                    line width=1.5pt,
                    mark size=2pt
                    ]
                    coordinates {
                    (0.8, 0.3764)
                    (0.85, 0.3735)
                    (0.9, 0.3657)
                    (0.95, 0.3617)
                    (1.0, 0.3599)
                    };
                    \addlegendentry{Species NMI}
                \addplot[
                    color=teal,
                    dotted,
                    mark options={solid},
                    mark=square*,
                    line width=1.5pt,
                    mark size=2pt
                    ]
                    coordinates {
                    (0.8, 0.7621)
                    (0.85, 0.7597)
                    (0.9, 0.7264)
                    (0.95, 0.7165)
                    (1.0, 0.7137)
                    };
                    \addlegendentry{Species RI}
                \end{axis}
                \end{tikzpicture}
    }
\subfigure[Clusterings on Fruit360]{
            \begin{tikzpicture}[font=\Large, scale=0.47]
                \begin{axis}[
                    legend cell align={left},
                    legend style={nodes={scale=1.0, transform shape}, at={(0.03,0.5)},anchor=west},
                    xlabel={$K$},
                    xtick pos=left,
                    tick label style={font=\large},
                    ylabel style={font=\large},
                    xtick={0.8, 0.85, 0.9, 0.95, 1.0},
                    xticklabels={$2$,$3$,$4$,$5$,$6$},
                    ytick={0.45, 0.55,0.65,0.75},
                    yticklabels={ $0.45$,$0.55$,$0.65$,$0.75$},
                    ymajorgrids=true,
                    grid style=dashed
                ]
                \addplot[
                    color=purple,
                    dotted,
                    mark options={solid},
                    mark=diamond*,
                    line width=1.5pt,
                    mark size=2pt
                    ]
                    coordinates {
                    (0.8, 0.4981)
                    (0.85, 0.4953)
                    (0.9, 0.4962)
                    (0.95, 0.4903)
                    (1.0, 0.4912)
                    };
                    \addlegendentry{Color NMI}
                \addplot[
                    color=blue,
                    dotted,
                    mark options={solid},
                    mark=*,
                    line width=1.5pt,
                    mark size=2pt
                    ]
                    coordinates {
                    (0.8, 0.7472)
                    (0.85, 0.7403)
                    (0.9, 0.7414)
                    (0.95, 0.7251)
                    (1.0, 0.7201)
                    };
                    \addlegendentry{Color RI}
                \addplot[
                    color=orange,
                    dotted,
                    mark options={solid},
                    mark=triangle*,
                    line width=1.5pt,
                    mark size=2pt
                    ]
                    coordinates {
                    (0.8, 0.5292)
                    (0.85, 0.5235)
                    (0.9, 0.5229)
                    (0.95, 0.5156)
                    (1.0, 0.5132)
                    };
                    \addlegendentry{Species NMI}
                \addplot[
                    color=teal,
                    dotted,
                    mark options={solid},
                    mark=square*,
                    line width=1.5pt,
                    mark size=2pt
                    ]
                    coordinates {
                    (0.8, 0.7703)
                    (0.85, 0.7698)
                    (0.9, 0.7715)
                    (0.95, 0.746)
                    (1.0, 0.7482)
                    };
                    \addlegendentry{Species RI}
                \end{axis}
                \end{tikzpicture}
    }
    \caption{Results of parameter sensitivity of $K$.}
    \label{fig:parameterK}
\end{figure}

\begin{figure}[t!]
\centering
\subfigure[Clusterings on Fruit]{
            \begin{tikzpicture}[font=\Large, scale=0.47]
                \begin{axis}[
                    legend cell align={left},
                    legend style={nodes={scale=1.0, transform shape}, at={(0.03,0.36)},anchor=west},
                    xlabel={$T$},
                    xtick pos=left,
                    tick label style={font=\large},
                    ylabel style={font=\large},
                    xtick={0.8, 0.85, 0.9, 0.95, 1.0},
                    xticklabels={$2$,$3$,$4$,$5$,$6$},
                    ytick={0.35, 0.55,0.75,0.95},
                    yticklabels={ $0.35$,$0.55$,$0.75$,$0.95$},
                    ymajorgrids=true,
                    grid style=dashed
                ]
                \addplot[
                    color=purple,
                    dotted,
                    mark options={solid},
                    mark=diamond*,
                    line width=1.5pt,
                    mark size=2pt
                    ]
                    coordinates {
                    (0.8, 0.8954)
                    (0.85, 0.8973)
                    (0.9, 0.8926)
                    (0.95, 0.8872)
                    (1.0, 0.8879)
                    };
                    \addlegendentry{Color NMI}
                \addplot[
                    color=blue,
                    dotted,
                    mark options={solid},
                    mark=*,
                    line width=1.5pt,
                    mark size=2pt
                    ]
                    coordinates {
                    (0.8, 0.9316)
                    (0.85, 0.9383)
                    (0.9, 0.931)
                    (0.95, 0.9273)
                    (1.0, 0.9255)
                    };
                    \addlegendentry{Color RI}
                \addplot[
                    color=orange,
                    dotted,
                    mark options={solid},
                    mark=triangle*,
                    line width=1.5pt,
                    mark size=2pt
                    ]
                    coordinates {
                    (0.8, 0.3751)
                    (0.85, 0.3764)
                    (0.9, 0.3715)
                    (0.95, 0.3695)
                    (1.0, 0.3626)
                    };
                    \addlegendentry{Species NMI}
                \addplot[
                    color=teal,
                    dotted,
                    mark options={solid},
                    mark=square*,
                    line width=1.5pt,
                    mark size=2pt
                    ]
                    coordinates {
                    (0.8, 0.7584)
                    (0.85, 0.7621)
                    (0.9, 0.7527)
                    (0.95, 0.7489)
                    (1.0, 0.7417)
                    };
                    \addlegendentry{Species RI}
                \end{axis}
                \end{tikzpicture}
    }
\subfigure[Clusterings on Fruit360]{
            \begin{tikzpicture}[font=\Large, scale=0.47]
                \begin{axis}[
                    legend cell align={left},
                    legend style={nodes={scale=1.0, transform shape}, at={(0.03,0.5)},anchor=west},
                    xlabel={$T$},
                    xtick pos=left,
                    tick label style={font=\large},
                    ylabel style={font=\large},
                    xtick={0.8, 0.85, 0.9, 0.95, 1.0},
                    xticklabels={$2$,$3$,$4$,$5$,$6$},
                    ytick={0.48, 0.58,0.68,0.78},
                    yticklabels={ $0.48$,$0.58$,$0.68$,$0.78$},
                    ymajorgrids=true,
                    grid style=dashed
                ]
                \addplot[
                    color=purple,
                    dotted,
                    mark options={solid},
                    mark=diamond*,
                    line width=1.5pt,
                    mark size=2pt
                    ]
                    coordinates {
                    (0.8, 0.4963)
                    (0.85, 0.4981)
                    (0.9, 0.4955)
                    (0.95, 0.4927)
                    (1.0, 0.4922)
                    };
                    \addlegendentry{Color NMI}
                \addplot[
                    color=blue,
                    dotted,
                    mark options={solid},
                    mark=*,
                    line width=1.5pt,
                    mark size=2pt
                    ]
                    coordinates {
                    (0.8, 0.7428)
                    (0.85, 0.7472)
                    (0.9, 0.7435)
                    (0.95, 0.7437)
                    (1.0, 0.7419)
                    };
                    \addlegendentry{Color RI}
                \addplot[
                    color=orange,
                    dotted,
                    mark options={solid},
                    mark=triangle*,
                    line width=1.5pt,
                    mark size=2pt
                    ]
                    coordinates {
                    (0.8, 0.5289)
                    (0.85, 0.5292)
                    (0.9, 0.5271)
                    (0.95, 0.5243)
                    (1.0, 0.5221)
                    };
                    \addlegendentry{Species NMI}
                \addplot[
                    color=teal,
                    dotted,
                    mark options={solid},
                    mark=square*,
                    line width=1.5pt,
                    mark size=2pt
                    ]
                    coordinates {
                    (0.8, 0.7688)
                    (0.85, 0.7703)
                    (0.9, 0.7673)
                    (0.95, 0.7596)
                    (1.0, 0.7559)
                    };
                    \addlegendentry{Species RI}
                \end{axis}
                \end{tikzpicture}
    }
    \caption{Results of parameter sensitivity of $T$.}
    \label{fig:paramterT}
    \vspace{-0.4cm}
\end{figure}
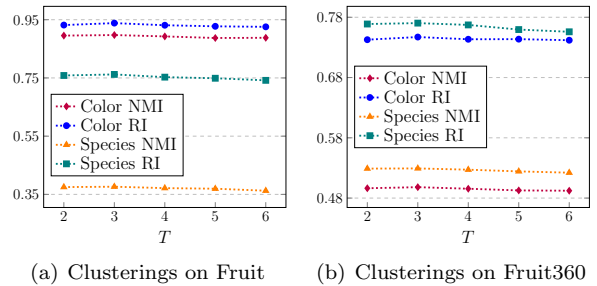

We further investigate the effect of the number of disentangled representations $K$ and the number of clusters $T$ of \sysname{}.
We change $K$ and $T$ from 2 to 6. The results of \sysname{} on Fruit and Fruit360 datasets under varying $K$ and $T$ are shown in Fig.~\ref{fig:parameterK} and Fig.~\ref{fig:paramterT}, respectively.
Regarding parameter $K$, \sysname{} performs best when $K=2$ for both Fruit and Fruit360 datasets, and the performance decreases as $K$ increases, indicating that $K=2$ is the optimal choice for the datasets. Specifically, for the color clustering of the Fruit dataset, performance experiences a significant downturn when $K$ reaches 5 or 6. This can be attributed to the introduction of excessive noise with an overly large $K$, thereby adversely affecting performance. Furthermore, for the parameter $T$, the performance of \sysname{} initially improves before declining as $T$ augments, with $T=3$ generating the best outcomes. These findings imply that an appropriate selection of the $K$ or $T$ value can enhance the effectiveness of \sysname{}.

\subsection{Efficiency Analysis}
\begin{figure}[t]
    \centering
    \includegraphics[width=0.4\textwidth]{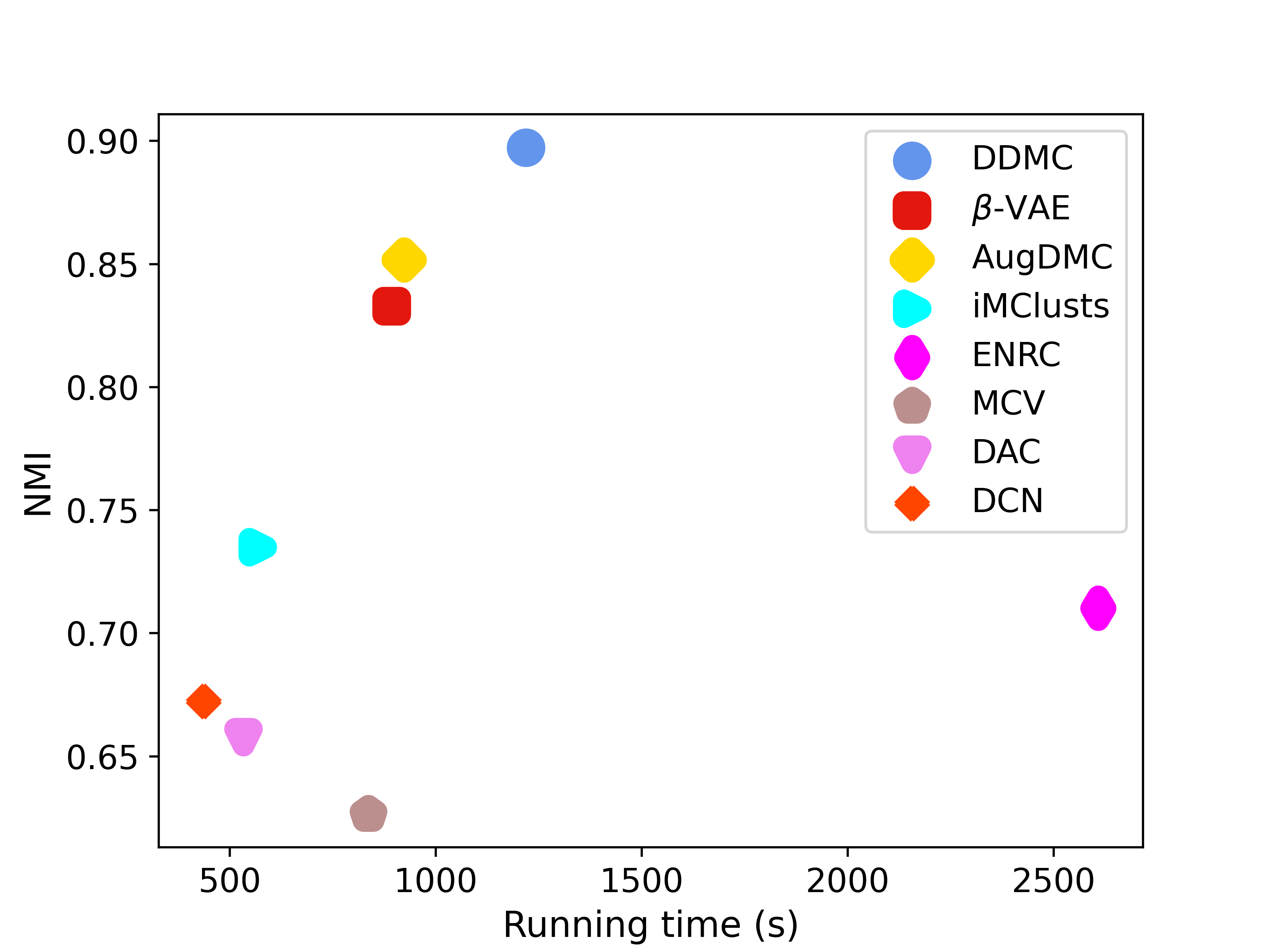}
    \caption{Performance v.s. the running time (s) on Fruit dataset.}
    \label{fig:time}
\end{figure}

Here we analyze the efficiency of the deep multiple clustering methods. The experiments are conducted on a server with a GPU GeForece RTX 2080Ti. We test the running time on Fruit dataset and the running time and related performance of color clustering are shown in Fig.~\ref{fig:time}. The two fastest methods are DAC and DCN, that is because both of them are single clustering learning methods and their module structure are simpler than the other multiple clustering methods. For the other methods, we can find methods with longer running time usually has better performance. \sysname{} can learn more effective image representations with acceptable efficiency.

\section{Conclusion}

In this paper, we present \sysname{}, a novel Dual-Disentangled deep Multiple Clustering method that leverages disentangled representations for multiple clustering. \sysname{} employs coarse-grained and fine-grained disentangled representations to reveal and disentangle the latent factors hidden in data. In addition, it incorporates cluster assignment module to further enhance the effectiveness and robustness of multiple clusterings in cluster-level performance. Furthermore, we formulate our method as a variational Expectation-Maximization framework and derive the Evidence Lower Bound of the fine-grained disentanglement. Extensive experiments on seven benchmark datasets demonstrate that \sysname{} attains state-of-the-art performance in terms of multiple clustering performance and each individual clustering's performance.
In future work, we intend to extend our methodology to manage more complex data types and scenarios, such as multi-modal data. Furthermore, compared to baseline methods, the proposed method is more costly although the performance has been improved. Improving the efficiency will be another interesting future direction.

\section{Acknowledgement}

This research is supported in part by Advata Gift funding. All opinions, findings, conclusions and recommendations in this paper are those of the author and do not necessarily reflect the views of
the funding agencies.
\bibliography{ref}
\bibliographystyle{siamplain}

\end{document}